\documentclass{article}

\usepackage{PRIMEarxiv}

\usepackage{natbib} % author-year citation commands (\citet, \citep)
\usepackage[utf8]{inputenc} % allow utf-8 input
\usepackage{amsmath}
\usepackage[T1]{fontenc} % use 8-bit T1 fonts
\usepackage{textgreek} % support Greek letters in text
\usepackage{hyperref} % hyperlinks
\usepackage{url} % simple URL typesetting
\usepackage{booktabs} % professional-quality tables
\usepackage{tabularx}
\usepackage{array}
\usepackage{amsfonts} % blackboard math symbols
\usepackage{nicefrac} % compact symbols for 1/2, etc.
\usepackage{microtype} % microtypography
\usepackage{multirow}
\usepackage{fancyhdr} % header
\usepackage{graphicx} % graphics
\graphicspath{{figures/}} % figures generated by evaluation scripts

\usepackage{flafter} % a float is placed only after its first reference

\title{Sub-Billion, Super-Frontier: Small Language Models Rival Zero-Shot Frontier LLMs on General and Literary Relation Extraction}

\author{
 Despina Christou \\
 School of Informatics,\\
 Aristotle University of Thessaloniki,\\
 54124, Greece \\
 \texttt{christoud@csd.auth.gr} \\
 \And
 Grigorios Tsoumakas \\
 School of Informatics,\\
 Aristotle University of Thessaloniki,\\
 54124, Greece \\
 Archimedes, Athena Research Center, Greece, \\
 \texttt{greg@csd.auth.gr} \\
}

\begin{document}
\maketitle

\begin{abstract}
Large language models (LLMs) achieve strong relation extraction (RE), but their computational demands and reliance on proprietary APIs limit deployment in resource-constrained or privacy-sensitive settings. We ask how far small language models (SLMs) can close this gap across general-domain and literary text, where implicit semantics and complex narrative structure pose additional challenges. We evaluate five models ranging from 360M to 3B parameters under three domain-composition regimes and two prompt-conditioned tuning styles (30 configurations), and compare them with zero-shot frontier LLMs and a discriminative RoBERTa baseline. Across nine benchmarks, using positive-class micro-F1, the best sub-billion model, Qwen2.5-0.5B fine-tuned on pooled general-domain data, achieves a general-domain average of 0.83, compared with 0.69 for GPT-5.4 and 0.66 for Claude Sonnet~4.6 under the same minimal zero-shot protocol. This comparison does not imply that SLMs are intrinsically stronger; rather, it shows that targeted task adaptation enables 4-bit models deployable on a single consumer GPU to outperform general-purpose frontier systems under this protocol. An in-domain RoBERTa baseline also exceeds both frontier models, indicating that the advantage stems from task adaptation rather than generative decoding. On literary RE, tuned SLMs lead GPT-5.4 by about 8 F1 points on the human-annotated Biographical benchmark (0.92 vs.\ 0.83) and by more than 25 points on the two-benchmark literary average (0.833 vs.\ 0.578), with the largest margin on the GPT-4o-annotated PG-Fiction corpus. A single-model domain-adaptive pretraining case study yields no practically meaningful improvement over supervised fine-tuning, while the cleanest same-generation, within-family scale comparison shows only marginal gains. These results suggest that, when task-specific data are available, compact task-adapted models can provide accurate, private, and hardware-efficient RE.
\end{abstract}

% keywords can be removed

\section{Introduction}
\label{sec:introduction}
Relation Extraction (RE), the task of identifying and classifying semantic relationships between entities in text, serves as a cornerstone for numerous downstream natural language processing (NLP) applications. High-quality RE supports knowledge graph construction \cite{carlson2010toward, dong2014knowledge}, structured search, question answering systems \cite{bordes2015large}, and broader text understanding \cite{zhao2024comprehensive}. 
Large language models (LLMs) have significantly advanced the state of the art in RE as well as in a wide range of NLP tasks, especially in zero-shot and few-shot settings. Frontier proprietary models such as ChatGPT \cite{openai2023gpt}, Claude \cite{anthropic2024claude3}, and Gemini \cite{team2024gemini, google2025gemini2.5}, alongside open-weight systems including Llama \cite{touvron2023llama}, Qwen \cite{yang2024qwen2}, and DeepSeek \cite{liu2024deepseek}, now exhibit strong relational reasoning and extraction capabilities
\cite{wadhwa2023revisiting, li2023revisiting}.

However, these gains come with significant computational, financial, and infrastructure costs. Models with tens or hundreds of billions of parameters require extensive compute for both training and inference, restricting deployment to settings with efficient hardware resources \cite{strubell2020energy, bender2021dangers}. Their large size also creates challenges for accessibility, sustainability, privacy, and adaptation, especially for proprietary models that are available only through APIs. As a result, many RE pipelines remain dependent on large-scale systems whose resource requirements limit wider and more equitable adoption.

These constraints have renewed interest in compact and efficient language models, especially those below one billion parameters \cite{lu2025demystifying, guo2025slms}. Such models promise faster inference, lower memory consumption, and compatibility with edge devices or privacy critical environments. In the meantime, sub-billion parameter models still lag behind their larger counterparts on complex reasoning tasks. This gap is particularly important for RE in domains with rich narrative structure, implicit semantics, figurative expressions, or long-range dependencies, phenomena common in literary texts \cite{bamman2019annotated, christou2024arf}. Despite rapid progress on scaling LLMs \cite{kaplan2020scaling, hoffmann2022training}, comparatively little work examines how far smaller models can be pushed toward frontier performance, especially under strict efficiency constraints that are critical for low-resource and on-device usage.

At the same time, recent studies suggest that compact models can be improved through better data and training strategies, such as domain-adaptive pretraining (DAPT) \cite{gururangan2020dont}, domain-specific data mixtures
\cite{gunasekar2023textbooks, allal2025smollm2}, and few-shot prompting \cite{brown2020language, schick2021exploiting}. These methods have shown benefits in tasks such as summarization, classification, and information extraction \cite{gururangan2020dont, wadhwa2023revisiting}. However, their combined effect on RE remains less well-studied, especially in specialized domains such as fiction \cite{christou2024arf}. It is therefore unclear how much systematic, task-focused optimization can reduce the performance gap between small models and frontier LLMs \cite{guo2025slms, bairi2025small}.

In this work, we examine whether small language models (SLMs), ranging from sub-billion to 3B parameters, can achieve strong RE performance on both general-domain and literary text. We evaluate five base models from three architectural families, each fine-tuned using three domain-composition settings and two prompt-based tuning styles, resulting in 30 tuned configurations. We further conduct a focused case study on domain-adaptive pretraining (DAPT) for literary RE and benchmark our SLMs with frontier proprietary LLMs. Across nine RE benchmarks, our best sub-billion-parameter model outperforms frontier LLMs on general-domain RE. For literary RE, our best 3B models exceed frontier LLMs by more than 25 F1 points while requiring only a fraction of the computation and storage; a targeted DAPT case study (one model, one corpus) indicates that, in this setting, the advantage stems from supervised task adaptation: continued pretraining on LitBank adds no practically meaningful gain. These findings show that targeted training strategies and data-centered optimization can make small models competitive with much larger systems. This offers a practical path toward more accessible RE systems that can be deployed in low-resource settings, support privacy-sensitive deployment by running locally rather than through third-party APIs, and extend coverage to domains underserved by current large-model pipelines.

Our key contributions are fourfold:
\begin{enumerate}
 \item \textbf{Systematic SLM evaluation for RE}: We conduct, to our knowledge, the most comprehensive controlled study to date of domain-composition and prompt-conditioned tuning for small-model relation extraction, spanning five base models from 360M to 3B parameters, three domain-composition tuning regimes, and two prompt-conditioned tuning styles, yielding 30 tuned configurations whose best instances surpass frontier proprietary LLMs on both general and literary benchmarks.
 \item \textbf{Multi-domain and cross-scale analysis}: Our evaluation suite spans nine RE benchmarks covering general-domain corpora (news, Wikipedia, web text) and literary texts, with each domain specialist evaluated on its in-domain datasets and the mixed-domain models on all nine, providing a controlled comparison of multi-domain learning, domain-balanced (mixed-domain) tuning, and specialist-versus-generalist coverage under consistent experimental conditions. Because the specialists are not evaluated outside their training domain, we make no claim about out-of-domain transfer or cross-domain generalization.
 \item \textbf{Analysis of training strategies and domain adaptation}: We quantify how mixed-domain tuning and prompt-conditioned supervision narrow the gap between small and frontier models, and, through a controlled domain-adaptive pretraining (DAPT) case study, show that the literary-domain gap is closed by supervised task adaptation rather than by continued in-domain pretraining.
 \item \textbf{Efficiency-oriented assessment}: We evaluate the performance--efficiency trade-off in terms of parameters, disaggregated deployment footprint, estimated single-example inference latency on consumer hardware (Section~\ref{subsec:implementation}), and normalized metrics (F1 per billion parameters), demonstrating that strong RE performance can be achieved with models deployable on a single consumer GPU or even a CPU.
\end{enumerate}

To support reproducibility and further research, we will publicly release our best-performing checkpoint, training configurations, datasets, and evaluation code.\footnote{Code, the best fine-tuned checkpoint, the processed datasets, and the collected frontier-model outputs: \url{https://github.com/DespinaChristou/compact-relex}. See Section~\ref{sec:data_code_availability} for further details.}

The remainder of this paper is organized as follows: Section 2 reviews related work. Section 3 describes our experimental design, including the task formulation, datasets, models, training regimes, and evaluation protocol. Section 4 presents results and analyses. Section 5 concludes with implications, limitations, and directions for future research.

% -----------------------------------------------------------------------------------------------------------------

\section{Related Work}
\label{sec:relatedWork}
We situate RE within five developments: the shift from discriminative encoders to generative and instruction-following models; the rise of efficient sub-billion SLMs and model compression; data-centric and few-shot supervision for low-resource RE; domain adaptation for literary text; and the sustainability/democratization agenda. The gap we address is the lack of systematic study of sub-billion, quantized \emph{generative} models for RE that are competitive across both general and literary domains.

\subsection{From Discriminative Encoders to Generative Reasoners}
\label{subsec:relatedWork/REParadigms}
% The paradigm shift in Relation Extraction
Classical neural RE used discriminative models that predicted relation labels from fixed schemas, from convolutional \cite{zeng2014relation} and recurrent or tree-based architectures \cite{miwa2016end} to transformer encoders: BERT \cite{devlin2019bert} and its derivatives SpanBERT \cite{joshi2020spanbert}, LUKE \cite{yamada2020luke}, and DeBERTa \cite{he2021deberta}, which offered richer contextualized representations for tasks like TACRED, ACE, and SemEval-2010 \cite{zhao2024comprehensive}. Span-based formulations that embed entity mentions and predict their relation \cite{eberts2020span, zhong2021pure} became a de facto standard, and encoder-based models remain strong baselines for NER and RE \cite{ruder2025encoder}.

These encoders have limitations that motivate more flexible approaches: they are tied to fixed ontologies, so adding or redefining relations requires retraining, costly in evolving domains \cite{gururangan2020dont, zhao2024comprehensive}; they exhibit a ``discriminative bottleneck,'' relying on surface patterns and confusing semantically close relations (e.g.\ birthplace vs.\ residence) in zero-shot or compositional settings \cite{wadhwa2023revisiting, zhao2024comprehensive}; and they need large labeled datasets, overfitting in low-resource settings rather than learning robust abstractions \cite{wadhwa2023revisiting}.

This motivated a pivot to generative information extraction, which reframes RE as sequence-to-sequence generation of linearized triples. \citet{paolini2021structured} cast diverse structured-prediction tasks as text-to-text generation with T5-style models \cite{raffel2020exploring}, and systems such as REBEL \cite{huguetcabot2021rebel} produce subject-relation-object sequences directly. Surveys \cite{huguetcabot2025generative, li2023survey, zhao2024comprehensive} note that generative IE can emit multiple relations per span, adapt to novel relation descriptions, and unify extraction tasks, but introduces hallucinated relations, malformed outputs, and difficulty enforcing schema constraints.

Hybrid methods bridge the two: generative models produce rationales (reasoning traces) that train smaller discriminative models or guide schema-faithful decoding \cite{li2025bridging, wadhwa2023revisiting}, but most target mid- to large-scale models. We instead ask how much generative reasoning sub-billion models optimized for RE can retain.

\subsection{LLMs for Relation Extraction}
\label{subsec:relatedWork/LLMs}
Frontier LLMs moved RE toward zero- and few-shot in-context learning (ICL). Proprietary models, GPT-3 \cite{brown2020language}, GPT-4 \cite{openai2023gpt}, Claude \cite{anthropic2024claude3}, and Gemini \cite{team2024gemini}, and open-weight families, LLaMA \cite{touvron2023llama}, Qwen \cite{bai2023qwen}, DeepSeek \cite{liu2024deepseek}, and Mistral \cite{jiang2023mistral}, perform RE without fine-tuning via prompting and a few exemplars \cite{brown2020language, schick2021exploiting}. SumAsk prompting recasts RE as question answering and lets zero-shot ChatGPT rival supervised baselines on TACRED \cite{li2023revisiting}, in-context example retrieval improves few-shot RE with GPT-3.5/4 \cite{wan2023gptre}, and schema-tailored prompt formats further narrow the gap to supervised methods \cite{wan2024grasping}.

Instruction tuning improves generalization to unseen IE tasks \cite{wei2021finetuned, ouyang2022training}, and chain-of-thought prompting \cite{wei2022chain} helps multi-hop and document-level RE \cite{arora2023multi, ma2023large, tao2024graphical}. Encoding annotation guidelines in the prompt enables strong zero-shot structured extraction across NER, RE, and event extraction \cite{sainz2024gollie}, while decoder-level constraints such as grammar-constrained decoding \cite{dagdelen2025grammar} guarantee well-formed outputs. We do \emph{not} use decoder-level constraints; instead, one of our two prompting conditions enumerates the allowed labels in the system prompt (\emph{schema-enumerated prompting}; Section~\ref{subsec:implementation}, Appendix~\ref{app:constrained_vs_open}), an advisory signal that does not by itself prevent out-of-schema outputs.

Yet deploying frontier LLMs for RE is expensive, latency-sensitive, and often gated behind proprietary APIs with usage and privacy concerns \cite{strubell2020energy, patterson2021carbon, bender2021dangers}, and both conventional and LLM-based methods struggle in low-resource multilingual settings \cite{jinensibieke2024multilingual, ali2025multilingual}. Retrieval-augmented fine-tuned LLMs mitigate some issues on TACRED \cite{efeoglu2024rag4re, efeoglu2025rag4re} but still assume large models. We diverge from the ``scaling-is-all-you-need'' view \cite{kaplan2020scaling, hoffmann2022training}, asking how far strategically optimized small models can go as an accessible alternative for robust RE.

\subsection{Small Language Models for Efficient Information Extraction}
\label{subsec:relatedWork/SLMs}
The rapid growth of frontier LLMs has been accompanied by sustained research into compact models that preserve performance while dramatically reducing compute and memory demands. Early work in efficient NLP focused primarily on encoder-based compression and architectural refinement: DistilBERT \cite{sanh2019distilbert} and TinyBERT \cite{jiao2020tinybert} applied knowledge distillation to reduce parameter counts with modest accuracy loss, ALBERT \cite{lanalbert} introduced parameter sharing to shrink memory footprints, and ELECTRA \cite{clarkelectra} proposed replaced-token detection to improve sample efficiency.

Although classical scaling laws predict predictable gains from increased model size, data, and compute \cite{kaplan2020scaling}, more recent studies demonstrate that with curated data and task-aware training, very small models can approach the performance of larger ones on targeted benchmarks \cite{bairi2025small, li2023survey}. This shift has fueled interest in small language models (SLMs), particularly sub-billion-parameter systems optimized explicitly for efficiency and deployment rather than sheer scale.

Recent decoder-only SLMs underline this trend. Qwen2-0.5B and its successors \cite{yang2024qwen2, qwen2025qwen2.5} show that training on massive corpora can compress substantial knowledge into compact models. Qwen2-0.5B Instruct achieves strong performance on reasoning and coding benchmarks and reliably generates syntactically valid structured outputs such as JSON, making it attractive for information extraction tasks where output correctness matters \cite{qwen2025qwen2.5}. Similarly, TinyLlama-1.1B demonstrates that terascale pretraining enables robust general-purpose performance on consumer hardware \cite{zhang2024tinyllama}. Data quality also plays a decisive role: the Phi series illustrates how curated ``textbook-quality'' synthetic reasoning data allow small models such as Phi-1 \cite{gunasekar2023textbooks}, Phi-3 Mini \cite{abdin2024phi3}, and early versions of Phi-4~\cite{abdin2024phi4} to approach or surpass substantially larger models on GSM8K and HumanEval. Google's Gemma family \cite{team2024gemma} similarly targets practical performance at moderate scale.

Architectural innovations specifically designed for edge deployment magnify these gains. MobileLLM and MobileLLM-Pro emphasize depth over width, embedding reuse, and optimized attention kernels, enabling real-time inference with long context windows on-device \cite{mobilellm2024}. These findings support the view that deeper, thinner networks often exhibit stronger reasoning behavior at fixed parameter budgets.

Lu et al.~\cite{lu2025demystifying} present the most comprehensive study of SLMs to date, benchmarking over 60 publicly available models across reasoning, code, and general tasks. Their findings confirm that state-of-the-art SLMs can outperform 7B models on general tasks but reveal that in-context learning capabilities remain limited at small scales, and that significant optimization potential exists through task-specific routing and hardware co-design. These observations motivate our approach of task-specific fine-tuning rather than relying on zero-shot ICL with small models.

Beyond model architecture, adaptation efficiency is crucial. Parameter-Efficient Fine-Tuning (PEFT) strategies such as adapter layers \cite{houlsby2019parameter}, LoRA \cite{hu2021lora}, and QLoRA \cite{dettmers2023qlora} update only small task-specific modules, making fine-tuning feasible on limited hardware. Surveys by Diaz-Garc\'{i}a et al.~\cite{diazgarcia2025peft} show that PEFT often recovers most of the performance of full fine-tuning across information extraction tasks, although SLMs still lag frontier LLMs in complex reasoning under zero-shot conditions. Despite this progress, relation extraction remains under-studied in the SLM literature: small models are typically included as baselines rather than optimized systems, and few studies investigate the combined effects of domain adaptation, data composition, and quantization-aware fine-tuning for generative RE. Compression is the key deployment enabler: large models tolerate 4-bit post-training quantization with little loss, whereas SLMs are more fragile, encoding information in a few high-magnitude outlier channels \cite{liu2025efficientllm, wang2025slm}; surveys of quantization and pruning catalog these trade-offs \cite{chen2024quantization, gholami2024survey, krishnamoorthi2018quantizing, zhao2024mixed, frantar2023sparsegpt, xu2025pruning}, and QLoRA \cite{dettmers2023qlora} sidesteps them by quantizing only the frozen base weights (4-bit NormalFloat) and training low-rank adapters, which suits our 24\,GB consumer-GPU setting.

Notable exceptions include encoder-based zero-shot systems such as GLiNER \cite{knowledgator2025gliner} and its relation extraction extension GLiREL \cite{boylan2025glirel}, which achieve strong results on FewRel and WikiZSL by framing RE as a matching problem in a shared latent space. A document-level variant, GLiDRE \cite{glidre2025}, has recently extended this approach to cross-sentence relations. However, these encoder-based systems are designed for span classification rather than open-ended structured generation and do not follow the decoder-only generative paradigm that our work targets. Recent evaluations contrasting SLMs and LLMs for RE and summarization \cite{jinensibieke2024multilingual, guo2025slms} further confirm the gap we aim to close: small generative models that are systematically optimized for RE through domain composition, prompt conditioning, and domain-adaptive pretraining.

\subsection{Synthetic Data and Few-Shot Strategies for Low-Resource RE}
\label{subsec:relatedWork/syntheticData}

The scarcity of labeled RE data, especially in specialized domains, has driven a shift from model-centric to data-centric methods. Distant supervision aligns knowledge-base triples with text \cite{mintz2009distant}, with later denoising via multi-instance learning \cite{lin2016neural} and label-aware embeddings \cite{christou2021improving}, but remains limited by knowledge-base coverage and label noise. LLMs now enable synthetic supervision: teacher models generate labeled examples and rationales conditioned on a target schema \cite{xu2023s2ynre, xu2024making, feng2024synthetic, jin2025clinical}, and chain-of-thought distillation trains students on teacher rationales, \citet{wadhwa2023revisiting} show Flan-T5 trained on GPT rationales beats zero-shot GPT-3.5 on TACRED despite far fewer parameters. Over-reliance on homogeneous synthetic data risks ``synthetic collapse,'' mitigated by multiple teachers or preference optimization, with mixed evidence on its efficiency benefits \cite{ding2025synthetic, gholami2023does}. In parallel, in-context learning frames low-resource RE as prompting rather than parameter updating \cite{brown2020language, min2022rethinking}, but in-context ability is strongly scale-dependent and weak at sub-billion scale.

Our work takes a complementary data-centric route: rather than generating synthetic examples, we use domain-composition tuning (mixing general and literary RE data in controlled proportions), domain-adaptive pretraining on unannotated literary text, and prompt-conditioned supervision with curated few-shot demonstrations, combined with QLoRA fine-tuning to remain within consumer-hardware limits \cite{guo2025slms}.

\subsection{Domain Adaptation in Literary Relation Extraction}
\label{subsec:relatedWork/domainAdaptation}

Domain shift remains a major obstacle to robust RE. General-purpose pretrained models struggle when confronted with specialized vocabulary, domain-specific entity types, or atypical discourse structures. Domain-adaptive pretraining (DAPT) and task-adaptive pretraining (TAPT) have proven effective at mitigating such shifts by continuing language model pretraining on in-domain or task-specific unlabeled text \cite{gururangan2020dont}. SciBERT \cite{beltagy2019scibert} and BioBERT \cite{lee2020biobert} are canonical examples that significantly outperform vanilla BERT on scientific and biomedical IE tasks, including relation extraction. More recent practitioner guides emphasize that even modest additional pretraining on domain-specific corpora can yield large downstream gains for custom applications \cite{gururangan2020dont}.

Literary and narrative domains pose unique challenges that go beyond vocabulary. Relations between characters are often implicit, evolve gradually, and are mediated by narrative voice and stylistic devices. Early computational literary studies focused on character identification, quotation attribution, and social network extraction from novels \cite{elson2010extracting, bamman2014bayesian}. Work such as \cite{chaturvedi2016modeling} modeled the evolution of character relations across narratives, showing that antagonistic or supportive relationships can shift through sequences of events. Evaluations of standard NER and RE tools on fiction have highlighted substantial drops in performance and difficulties with coreference, indirect speech, and culturally specific expressions \cite{jaschke2021named}.

Recent research began targeting literary RE more directly. Christou and Tsoumakas~\cite{christou2021literary, christou2024arf} explored relation extraction in Greek and English literary texts, noting that conventional RE models struggle with indirect and implicit relations, as well as with long-range dependencies that span chapters. To alleviate annotation bottlenecks, they proposed the Artificial Relationships in Fiction (ARF) dataset, which uses GPT-4 to synthetically annotate literary texts with a rich ontology of social and narrative relations. Parallel work on context-aware implicit relation discovery in multi-event chains underscores the importance of modeling discourse structure and context to capture latent relational semantics \cite{zhao2025implicit}. Studies on quotation attribution and narrative generation also reveal that current LLMs tend to favor stable, stereotypical narrative patterns, potentially biasing downstream RE systems if synthetic stories are naively used as training data \cite{shaw2024quotation}.

In this context, our work is one of the first to examine whether small generative models (360M--3B parameters), equipped with DAPT on literary corpora and domain-composition tuning strategies, can tackle the implicit and long-range nature of literary relations competitively with frontier LLMs, while remaining deployable on consumer-grade hardware.

\subsection{Sustainability, Energy, and Democratization}
\label{subsec:relatedWork/sustainability}

Concerns about the environmental and social costs of large models have made efficiency a first-class objective. Schwartz et al.~\cite{schwartz2020green} distinguish accuracy-at-any-cost ``Red AI'' from efficiency-aware ``Green AI,'' and others document the carbon and concentration-of-power costs of ever-larger models \cite{strubell2020energy, bender2021dangers}. As inference dominates lifecycle energy for deployed models \cite{stromer2025inference, acl2025energy}, right-sizing models to the task is among the most impactful levers, task-aware selection alone could cut AI energy substantially \cite{smallissufficient2025}, and sub-billion models quantized to 4-bit on local CPUs or NPUs can reduce energy per token by an order of magnitude \cite{zhang2025quantization, singh2025power}. Such models also keep data on-device, easing the privacy and regulatory concerns of cloud APIs. We measure the performance--efficiency trade-off of 360M--3B generative RE models in this spirit, showing that domain-composition tuning and adaptation let compact models match or exceed frontier LLMs on RE.
% ======================================================

\section{Experimental Design}
\label{sec:experimental_design}

Our experimental design focuses on bridging the performance gap between SLMs ranging from 360M to 3B parameters and frontier models for RE across both general and complex literary domains. We combine task-specific prompt formulation, controlled domain-composition fine-tuning, quantized low-rank adaptation, and a targeted domain-adaptive pretraining (DAPT) case study. This section describes the task formulation, datasets, models, prompting strategy, training regimes, implementation details, and evaluation protocol.

\subsection{Task Formulation}
\label{subsec:task_formulation}

We formulate relation extraction as a text-generation task. Given an input sentence and two marked entities, a head entity and a tail entity, the model must generate the semantic relation holding between them from a predefined dataset-specific label set. Because the entity pair is provided rather than detected, this task is, strictly, sentence-level \emph{relation classification} given marked entity mentions, the standard formulation of benchmarks such as TACRED and SemEval-2010 Task~8, rather than open relation \emph{extraction}, which must additionally detect entities and the (mostly negative) candidate pairs; we use the term ``relation extraction'' throughout, following common usage in this literature. Several datasets additionally include a catch-all class, such as \textit{NA}, \textit{Other}, or \textit{none}, for entity pairs whose relation is not covered by the positive relation schema. This requires the model not only to select the correct relation type when one exists, but also to determine when no schema relation is expressed.

All datasets are converted into a common prompt-based format while preserving their original train, validation, and test splits whenever available. Evaluation is conducted on the official test split of each dataset. For datasets without an official validation split, we create one from the training data using stratified sampling to preserve the relation-label distribution.

\subsection{Datasets}
\label{subsec:datasets}

Our evaluation spans two domains: \textit{General}, covering news, web, Wikipedia, and document-level RE benchmarks, and \textit{Literature}, covering biographical and fiction-based relation extraction. Table~\ref{tab:datasets} summarizes the datasets used in this study.

\begin{table}[ht]
\centering
\footnotesize
\setlength{\tabcolsep}{5pt}
\begin{tabular}{l l l r r c}
\toprule
\textbf{Domain} & \textbf{Dataset} & \textbf{Reference} & \textbf{Samples} & \textbf{\#Rel.} & \textbf{\#Ent.\ Types} \\
\midrule
\multirow{7}{*}{General}
 & TACRED & \citep{zhang2017position} & 68,124 & 41 (+NA) & --$^\ddagger$ \\
 & SemEval-2010 Task 8 & \citep{hendrickx2010semeval} & 8,000 & 9 (+Other)$^\dagger$ & -- \\
 & CoNLL04 & \citep{roth2004linear} & 1,283 & 5 & 4 \\
 & NYT11 & \citep{hoffmann2011knowledge} & 94,222 & 24 & 4 \\
 & GIDS & \citep{jat2018improving} & 11,297 & 4 (+NA) & -- \\
 & Re-DocRED & \citep{tan2022revisiting} & 80,450 & 96 & 6 \\
 & REBEL & \citep{huguetcabot2021rebel} & 150,000/3.98M & 268$^\flat$ & -- \\
\midrule
\multirow{3}{*}{Literature}
 & Biographical & \citep{plum2022biographical} & 300,000/1.35M & 9 (+Other) & -- \\
 & PG-Fiction & \citep{christou2024arf} & 95,000 & 137 (+none)$^*$ & 11 \\
 & LitBank & \citep{bamman2019annotated} & -- & -- & -- \\
\bottomrule
\end{tabular}
\caption{Relation extraction datasets used in this study, grouped by domain. Counts and conventions are described in Section~\ref{subsec:datasets}.}
\label{tab:datasets}
\end{table}

Samples is the number of training instances used (the original corpus size follows the slash for the subsampled REBEL and Biographical); \#Rel.\ is the number of positive relation types, with a (+NA)/(+Other)/(+none) suffix marking a catch-all class for pairs with no schema relation; \#Ent.\ Types counts entity types exposed in the prompt. Evaluation always uses each dataset's official test split.

A few datasets warrant clarification. SemEval-2010 Task~8 defines nine direction-sensitive relations (18 directional labels plus \textit{Other}). REBEL is built from Wikidata properties with no fixed schema; we report the 268 relation types observed in our subsample. TACRED's fine-grained NER types are omitted so its prompts use the same mention-only format as the entity-type-free datasets. PG-Fiction contains 137 positive labels in our processed corpus, exceeding the 48-relation ARF ontology~\citep{christou2024arf} because the GPT-4o annotator emitted fine-grained subtypes; we therefore evaluate it under two inventories, reported as co-primary: the full 137-label corpus (conservative) and the canonical 48-relation ontology (mapping the 90 out-of-ontology labels, 2.3\% of positive instances, to the background class; Appendix~\ref{app:pgfiction48}). Entity-type annotations exist only for CoNLL04, NYT11, Re-DocRED, and PG-Fiction; LitBank carries no relation labels and is used only for domain-adaptive pretraining (Section~\ref{sec:dapt}).

\subsection{Models}
\label{subsec:base_models}
To study the trade-off between model size, domain adaptation, and RE performance, we evaluate five open-weight SLMs across two parameter scales and compare them against frontier proprietary LLMs:

\begin{itemize}
 \item \textbf{Sub-billion} (under 500M parameters): SmolLM2-360M-Instruct \cite{allal2025smollm2} and Qwen2.5-0.5B-Instruct \cite{qwen2025qwen2.5}.
 \item \textbf{Small} (3B parameters): SmolLM3-3B \cite{bakouch2025smollm3}, Qwen2.5-3B-Instruct \cite{qwen2025qwen2.5}, and Llama-3.2-3B-Instruct \cite{grattafiori2024llama3}.
 \item \textbf{Frontier LLMs}: GPT-5.4 and Claude Sonnet 4.6, evaluated as zero-shot reference baselines and accessed through their OpenRouter API model identifiers (Section~\ref{subsec:implementation}); capability and configuration details are documented in the providers' official model documentation and system cards. We additionally queried Gemini~2.5~Pro \cite{google2025gemini2.5}, but exclude it from the main comparison because it failed to produce schema-valid output (see Table~\ref{tab:general_frontier_llm_comparison}).
\end{itemize}

The SLMs were selected for open availability, instruction-tuned variants, compact disk footprint, and feasibility of fine-tuning and inference on consumer hardware. The model set also spans three families, SmolLM, Qwen, and Llama, allowing us to test whether observed trends are robust across different pretraining corpora, tokenizers, and instruction-tuning procedures. Frontier LLMs are included to contextualize the performance gap between compact task-adapted models and large proprietary systems.
% ------------------------------------------

\subsection{Prompt Design}
\label{sec:prompt_design}
Each relation instance is rendered as a natural-language prompt. We evaluate both zero-shot and few-shot prompt formats in order to measure how models perform under different amounts of task-specific context. To reduce template-specific bias, each instance is assigned one of ten instruction paraphrases uniformly at random. The same template pool is shared across all datasets and is listed in Appendix~\ref{app:example_prompts}.

\subsubsection{Zero-Shot Prompts}
In the zero-shot setting, the model receives the sentence, the head entity, and the tail entity, and must generate the relation label without any preceding demonstrations. For datasets that provide entity type information, types may also be included inline with the corresponding entity mentions, as described below. An example zero-shot prompt is provided in Appendix~\ref{app:zero_shot}.

\subsubsection{Entity Types}
Four datasets provide entity-type annotations (CoNLL04, NYT11, Re-DocRED, PG-Fiction), rendered inline with the mention, e.g.\ \textit{``Avinor [ORG]''}. Types are available for essentially all instances in the three general sets and ${\sim}78\%$ of PG-Fiction; at inference we include all available types, and the remaining PG-Fiction examples use mentions alone.

During fine-tuning, entity types are included stochastically as a form of input regularization. Specifically, for examples with entity type annotations, the prompt includes head and tail entity types with probability 80\% and omits them with probability 20\%. This \emph{type dropout} encourages the model to exploit type information when available while remaining robust for examples or datasets where type annotations are unavailable. The training setup is described further in Section~\ref{subsec:training}.

\subsubsection{Few-Shot Prompts}
In the few-shot setting, we prepend two demonstration examples to the query prompt. Demonstrations are stratified by relation class, with the two examples selected from different relation classes where possible, in order to provide diverse task context.

To prevent data leakage, demonstration selection depends on the split being rendered. For training prompts, demonstrations are drawn from the training split. For validation and test prompts, demonstrations are drawn exclusively from the training split, ensuring that evaluation instances never serve as demonstrations. Demonstrations follow the same entity-type rendering policy as the query they accompany. The few-shot prompt template is provided in Appendix~\ref{app:few_shot}.

\subsection{Training Regimes}
\label{subsec:training}

Our experimental design is structured to answer three key research questions: (1) What is the impact of few-shot examples during tuning? (2) How does a model tuned on a mixture of general and domain-specific data perform on a specialized domain? (3) Does domain-adaptive pre-training improve literary RE beyond supervised fine-tuning alone?

To address questions (1) and (2), we designed three core finetuning regimes, each tested in zero-shot and two-shot prompt-conditioned configurations. Combined with the five base SLMs, this yields 30 tuned configurations (5 models $\times$ 3 regimes $\times$ 2 prompt styles), enabling controlled analysis across model family, parameter scale, prompt format, and domain composition.

We distinguish two senses of ``shot'' that this design deliberately couples. The \emph{tuning shot} is the number of in-context demonstrations present in the prompts seen during fine-tuning (0 or 2); it indexes the 30 configurations above. The \emph{prompt shot} is the number of demonstrations supplied to a checkpoint at inference time. Unless stated otherwise, the two are \emph{matched}, a 0-shot-tuned model is evaluated with a 0-shot prompt and a 2-shot-tuned model with a 2-shot prompt, so a single ``0-shot''/``2-shot'' label denotes the matched training-and-inference pipeline. We additionally generate the off-diagonal case in which a 2-shot-tuned checkpoint is evaluated with a 0-shot prompt, which lets us separate the training-time and inference-time roles of demonstrations in Section~\ref{subsec:scale_prompt}.

The three core finetuning regimes are:
\begin{itemize}
\item \textbf{GenTune}: Model tuned only on the combined General domain datasets.
\item \textbf{LitTune}: Model tuned only on the combined Literature domain datasets.
\item \textbf{MixTune}: Model tuned on a balanced mixture of General and Literature datasets.
\end{itemize}

Source datasets are pooled at the example level and shuffled with seed $42$: GenTune pools the seven general benchmarks, LitTune the two literary ones, and MixTune is \emph{domain-balanced}, drawing equal numbers of general and literary examples (the larger pool subsampled to match). No per-source cap is applied within a domain, so each dataset contributes in proportion to its pool size (Table~\ref{tab:datasets}); the general mixture is thus dominated by REBEL, NYT11, Re-DocRED, and TACRED, with CoNLL04 and SemEval under 2\% each. Each run draws at most 200{,}000 examples (Table~\ref{tab:hyper_finetune}), preserving the ${\sim}$50/50 split for MixTune; the largest corpora (REBEL, Biographical) are themselves subsampled, and per-regime composition is reported in Appendix~\ref{app:dataset_stats} (Table~\ref{tab:train_composition}).

To address question (3), we conduct a focused DAPT case study on Llama-3.2-3B-Instruct, comparing DAPT-then-fine-tuned variants against the corresponding non-DAPT baselines. We select Llama-3.2-3B because it is the strongest model on general-domain RE and within a point of the best literary configuration, while behaving stably across all regimes; the marginally stronger literary model, SmolLM3-3B, is unstable in the 0-shot MixTune configuration, where its reasoning mode must be disabled at inference to emit a label (Section~\ref{sec:literary_re}), which would complicate the controlled DAPT comparison. This is treated as a targeted follow-up rather than a fourth full training regime, keeping the core experimental grid balanced and interpretable.

\subsection{Domain-Adaptive Pre-training}
\label{sec:dapt}
Given the substantial stylistic and semantic differences between general-purpose text, such as news or web text, and literary narratives, we investigate whether domain-adaptive pretraining (DAPT) before task-specific fine-tuning improves literary relation extraction. In this case study, the strongest-performing base model in the main experiments, Llama-3.2-3B-Instruct, undergoes continued pretraining with a causal language modeling objective on unannotated LitBank text~\cite{bamman2019annotated}, exposing the model to literary vocabulary, syntax, discourse patterns, and narrative structure before supervised RE tuning. The resulting DAPT checkpoint is then fine-tuned under LitTune and MixTune in the 0-shot configuration, yielding a 2$\times$2 comparison that crosses DAPT vs.\ no-DAPT with literature-only vs.\ mixed-domain supervision. We omit GenTune here because literary DAPT followed by general-domain-only fine-tuning would introduce a domain mismatch that confounds attribution; restricting the comparison to LitTune and MixTune isolates the effect of literary adaptation.

\subsection{Implementation Details}
\label{subsec:implementation}
All SLMs are fine-tuned using QLoRA \cite{dettmers2023qlora}, which combines 4-bit NormalFloat quantization of the base model weights with low-rank adaptation (LoRA) of attention and feed-forward projection layers. Specifically, we apply LoRA to all seven projection matrices (\texttt{q\_proj}, \texttt{k\_proj}, \texttt{v\_proj}, \texttt{o\_proj}, \texttt{gate\_proj}, \texttt{up\_proj}, \texttt{down\_proj}) with rank $r{=}64$ and scaling factor $\alpha{=}128$ for the 3B models, and proportionally smaller ranks for sub-billion models ($r{=}16$ for SmolLM2-360M, $r{=}32$ for Qwen2.5-0.5B); trainable parameters range from 8.7M to 121M (2.4--3.9\% of the backbone), with full per-model adapter and artifact sizes (base, 4-bit, FP32 adapter, merged, and GGUF) in Table~\ref{tab:artifact_sizes} (Appendix~\ref{app:hyperparameters}). We train for 2 epochs with an effective batch size of 8 (per-device batch size 4 with gradient accumulation over 2 steps), using the paged AdamW 8-bit optimizer with a learning rate of $1 \times 10^{-4}$, linear warmup over 3\% of steps, and weight decay of 0.01. The maximum sequence length is 1{,}024 tokens, and data subsampling and few-shot demonstration selection use a fixed random seed ($42$). Sequences are right-truncated at this $1{,}024$-token cap during training, with the gold label appended last; at inference, inputs are truncated only at each tokenizer's native context length ($8{,}192$--$131{,}072$ tokens). The resulting truncation is negligible, $0\%$ at inference and $0\%$ for seven of the nine datasets at training, and is reported by dataset, prompting condition, and tokenizer in Appendix~\ref{app:truncation}.

\paragraph{Model selection and seeds.} All design choices are fixed in advance and uniform across the 30 configurations, so no model is chosen on the basis of test performance: every run uses the same hyperparameters (Appendix~\ref{app:hyperparameters}), tuned for none individually, and we always evaluate the \emph{final} checkpoint after two epochs (\texttt{save\_total\_limit}{=}1, with no best-validation or early-stopping selection); the validation split is used only for loss monitoring during training, not for checkpoint or configuration selection. No configuration, prompt, or evaluation protocol was altered after inspecting test results: the two anomalous 0-shot configurations (SmolLM3-3B MixTune, Qwen2.5-3B GenTune) are reported under the same default protocol as every other run, and the reasoning-disabled recovery for SmolLM3-3B is disclosed separately as an explicitly post-hoc analysis in Section~\ref{sec:results}. Each configuration is trained once under a single seed ($42$). The bootstrap confidence intervals in Section~\ref{sec:statistical_significance} therefore quantify test-set sampling variance only and do \emph{not} capture training-time variability (QLoRA initialisation, data subsampling, demonstration selection, or optimisation stochasticity); accordingly, the near-tied top-of-table comparisons, notably the sub-billion-versus-best-3B-generalist result, are reported as such and are not claimed to be robust to reseeding.

At inference, SLMs decode with near-greedy settings (temperature $0.001$, top-$p{=}1.0$, no repetition penalty) and a 128-token generation budget. We compare two \emph{prompting} conditions, not decoder-level constraints, using the same checkpoints: \emph{generic prompting}, in which the system prompt does not enumerate labels, and \emph{schema-enumerated prompting}, in which the system prompt is augmented with the dataset's allowed label set. Schema enumeration is advisory only: the decoder remains unconstrained and can still emit out-of-schema labels, so this is schema-enumerated \emph{prompting} rather than grammar- or logit-constrained decoding. Unless otherwise noted, reported results refer to the \emph{schema-enumerated}
setting.

Frontier LLMs (GPT-5.4, Claude Sonnet 4.6) are queried zero-shot via the OpenRouter API (\texttt{openai/gpt-5.4}, \texttt{anthropic/claude-sonnet-4.6}) at temperature $0$ with a 64-token limit (non-binding, since labels are short). We set no reasoning-effort or routing parameters, so requests use each provider's default effort and routing, reflecting a default, low-overhead configuration rather than a reasoning-maximized one; as OpenRouter may route to different backends without a pinned provider, exact reproducibility is not guaranteed, so we release all collected frontier generations and the full request configuration, retry policy, failure counts, and dates in Appendix~\ref{app:frontier_prompts}.

All fine-tuning and SLM generation runs on a single NVIDIA RTX~4090 (24\,GB); throughout, ``consumer hardware'' means this class (a single RTX~4090 or an Intel Core i7-13700K CPU), without server-class or multi-GPU infrastructure. The stack is PyTorch~2.6 with \texttt{transformers}, \texttt{datasets}, \texttt{peft}, and \texttt{bitsandbytes}. The full 30-configuration grid took ${\sim}600$ GPU-hours (${\sim}16$\,h per sub-billion run, up to 22\,h per 3B run), plus ${\sim}50$ GPU-hours for the DAPT case study.

\paragraph{Latency protocol.}
The latencies in Table~\ref{tab:efficiency_tradeoffs} are \emph{estimated} single-example figures (batch size~1; a ${\sim}150$-token prompt with a ${\sim}5$-token completion), order-of-magnitude guides rather than benchmarked means, and enter no F1 computation. The GPU estimate uses the 4-bit (NF4) QLoRA checkpoint on the RTX~4090 via \texttt{transformers}/\texttt{bitsandbytes}; the CPU estimate uses the same model exported to \texttt{llama.cpp} (\texttt{Q4\_K\_M}) on the i7-13700K. F1 always uses the NF4 checkpoints.

QLoRA's 4-bit quantization enables training and inference up to 3B parameters within 24\,GB, and subsampling the largest datasets (REBEL, Biographical) keeps each run within a 24-hour window.
% ----------------------------------------------------------
\subsection{Evaluation Protocol and Metrics}
\label{subsec:evaluation}
The evaluation suite spans nine benchmarks (seven general-domain, two literary). Following the domain-specialization design (Table~\ref{tab:eval_matrix}), GenTune is evaluated on the seven general benchmarks, LitTune on the two literary ones, and MixTune on all nine. We report individual datasets, grouped domain averages, and an overall average, using dataset-macro averaging throughout so larger datasets do not dominate the ranking (the benchmarks differ substantially in size, ontology, and granularity): General Avg over the seven general benchmarks, Literature Avg over the two literary ones, and Overall Avg over all nine.

\paragraph{Output normalization.}
Because the models produce free-form text, each generation is normalized before scoring: we take the first line, collapse whitespace, lowercase, and strip surrounding quotes, applying no alias mapping (so labels keep their original delimiters, e.g.\ \textit{org:top\_members/employees}). A prediction is correct only on an exact match to the normalized gold label, so a well-formed but out-of-schema prediction counts as incorrect; the reported scores are therefore a conservative lower bound on performance.

\paragraph{Metrics.}
Our primary metric is \emph{positive-class micro-F1}: micro-averaged F1 over the positive relation types, excluding the catch-all/no-relation class, following the standard RE convention (e.g.\ TACRED). Because the catch-all dominates several benchmarks (78.6\% of TACRED, 52.8\% of Biographical; Appendix~\ref{app:dataset_stats}), counting it as an ordinary label would conflate RE with majority-class abstention; we exclude it from the averaged classes while still penalizing a positive prediction on a catch-all gold (false positive) and a catch-all prediction on a positive gold (false negative). We co-report \emph{positive-class macro-F1} (the unweighted mean of per-relation F1), which is more sensitive to the rare-relation tail and more informative for large-schema benchmarks such as PG-Fiction, for which we additionally report both metrics under the dataset's canonical 48-relation ontology (Appendix~\ref{app:pgfiction48}), mapping the 90 non-canonical labels to the background class. Since each instance receives a single label, all-class micro-F1 equals accuracy; we report this only as a secondary figure (labelled accuracy). Where an official scorer exists we also report the native metric for comparability (direction-aware macro-F1 for SemEval-2010 Task~8; micro-F1 excluding \textit{no\_relation} for TACRED); for GIDS, Re-DocRED, and REBEL our sentence-level setup differs from the native bag- or document-level evaluation, so those rows are not directly comparable to published leaderboards. Catch-all surface forms (\textit{NA}, \textit{Other}, \textit{none}, empty) are unified to one negative class for gold and predictions. Finally, two output-quality diagnostics that do not affect scoring characterize generation reliability: the \emph{schema-valid rate} (outputs matching a schema label after normalization) and the \emph{malformed rate} (empty or implausibly long outputs).

\paragraph{Statistical reliability.}
For all reported F1 scores we compute 95\% bootstrap confidence intervals with 10{,}000 iterations, and for key pairwise comparisons we apply paired bootstrap tests on positive-class F1~\cite{koehn2004statistical, efron1993introduction} rather than a McNemar test~\cite{mcnemar1947note}, which assumes paired binary outcomes (Section~\ref{sec:statistical_significance}). The full per-dataset positive-class F1 matrix for all 30 configurations is provided in Appendix~\ref{app:full_results}.

\paragraph{Follow-up analyses.}
Beyond the 30-model matrix, two follow-up analyses (not additional training regimes) round out the evaluation: a targeted DAPT study testing whether continued literary adaptation adds gains beyond supervised literature tuning, and a comparison of the strongest SLM configurations against frontier proprietary models on the general (Section~\ref{sec:results}) and literary (Section~\ref{sec:literary_re}) benchmarks.

\begin{table*}[t]
\centering
\scriptsize
\setlength{\tabcolsep}{4pt}
\renewcommand{\arraystretch}{1.1}
\begin{tabularx}{\columnwidth}{lX>{\raggedright\arraybackslash}X}
\toprule
\textbf{Factor} & \textbf{Values / Settings} & \textbf{Description} \\
\midrule
Base Models & SmolLM2-360M-Instruct; Qwen2.5-0.5B-Instruct; SmolLM3-3B; Qwen2.5-3B-Instruct; Llama-3.2-3B-Instruct & Base SLM families used for tuning. \\
Prompt-conditioned tuning styles & 0-shot; 2-shot & Prompt format used during fine-tuning. \\
Tuning regimes & GenTune; LitTune; MixTune & General-only, literature-only, or mixed-domain supervision. \\
General datasets & TACRED; SemEval-2010 Task 8; CoNLL04; NYT11; GIDS; Re-DocRED; REBEL & General-domain RE benchmarks. \\
Literature datasets & Biographical; PG-Fiction & Literature-domain RE benchmarks. \\
Main metric & Micro-F1 & Primary dataset-level performance measure. \\
Secondary metrics & Precision; Recall & Additional error-profile measures. \\
Aggregate reporting & General Avg; Literature Avg; Overall Avg & Dataset-macro averages across benchmark groups. \\
Generative robustness & Valid-schema rate; malformed output rate & Output-format reliability measures. \\
Extensions & Targeted DAPT (literature); Frontier LLM comparison (both domains) & Follow-up analyses beyond the 30-model grid. \\
\bottomrule
\end{tabularx}
\caption{Overview of the evaluation setup. The main study comprises 30 tuned SLMs obtained from 5 base models, 2 prompt-conditioned tuning styles, and 3 tuning regimes, evaluated over 9 relation extraction benchmarks grouped into General and Literature domains.}
\label{tab:eval_matrix}
\end{table*}

% ------------------------------------------------------------------------------

\section{Results}
\label{sec:results}
We organize the evaluation around three questions: (1)~how do tuned SLMs perform across general and literary domains under specialist versus mixed-domain supervision, (2)~how is performance shaped by model scale and prompt-conditioned supervision, and (3)~how far can literary RE be pushed through targeted adaptation? We complement the quantitative analysis with qualitative error study and an efficiency-oriented discussion motivated by the broader goal of democratizing relation extraction. Table~\ref{tab:eval_matrix} summarizes the experimental design (30 configurations = 5 models $\times$ 3 regimes $\times$ 2 prompt styles, as described in Section~\ref{subsec:training}).

\subsection{Multi-Domain Performance and Mixed-Domain Tuning}
We begin with the main comparative results across all 30 tuned SLMs. Our goal in this section is to determine which configurations perform best overall, which models are strongest within each domain, and whether mixed-domain supervision offers a more robust alternative to domain-specialized tuning.

\paragraph{Overall performance.}

Table~\ref{tab:main_summary_results} presents the main summary of results, reporting the General-domain and Literature-domain dataset-macro averages for every tuned configuration. By design (Section~\ref{subsec:training}), the domain specialists GenTune and LitTune are evaluated only on their own domain, while MixTune is evaluated on both; we therefore report performance per domain rather than as a single pooled average, which would mix in-domain and out-of-domain coverage across regimes and would not be comparable.

% Table 3
\begin{table*}[t]
\centering
\small
\setlength{\tabcolsep}{6pt}
\renewcommand{\arraystretch}{1.12}
\begin{tabularx}{\textwidth}{lllc>{\centering\arraybackslash}X>{\centering\arraybackslash}X}
\toprule
\textbf{Base Model} & \textbf{Params} & \textbf{Prompt Style} & \textbf{Tuning Regime} & \textbf{General Avg F1} & \textbf{Literature Avg F1} \\
\midrule

\multirow{6}{*}{SmolLM2-360M-Instruct}
& \multirow{6}{*}{360M} & 0-shot & GenTune & 0.527 & -- \\
& & 0-shot & LitTune & -- & 0.619 \\
& & 0-shot & MixTune & 0.555 & 0.549 \\
& & 2-shot & GenTune & 0.735 & -- \\
& & 2-shot & LitTune & -- & 0.756 \\
& & 2-shot & MixTune & 0.747 & 0.760 \\
\midrule

\multirow{6}{*}{Qwen2.5-0.5B-Instruct}
& \multirow{6}{*}{0.5B} & 0-shot & GenTune & 0.714 & -- \\
& & 0-shot & LitTune & -- & 0.780 \\
& & 0-shot & MixTune & 0.678 & 0.724 \\
& & 2-shot & GenTune & 0.828 & -- \\
& & 2-shot & LitTune & -- & 0.799 \\
& & 2-shot & MixTune & 0.810 & 0.773 \\
\midrule

\multirow{6}{*}{SmolLM3-3B}
& \multirow{6}{*}{3B} & 0-shot & GenTune & 0.819 & -- \\
& & 0-shot & LitTune & -- & \textbf{0.833} \\
& & 0-shot & MixTune & 0.000$^\dagger$ & 0.000$^\dagger$ \\
& & 2-shot & GenTune & 0.833 & -- \\
& & 2-shot & LitTune & -- & 0.806 \\
& & 2-shot & MixTune & 0.775 & 0.816 \\
\midrule

\multirow{6}{*}{Qwen2.5-3B-Instruct}
& \multirow{6}{*}{3B} & 0-shot & GenTune & 0.277$^\ddagger$ & -- \\
& & 0-shot & LitTune & -- & 0.822 \\
& & 0-shot & MixTune & 0.798 & 0.804 \\
& & 2-shot & GenTune & 0.824 & -- \\
& & 2-shot & LitTune & -- & 0.825 \\
& & 2-shot & MixTune & 0.793 & 0.809 \\
\midrule

\multirow{6}{*}{Llama-3.2-3B-Instruct}
& \multirow{6}{*}{3B} & 0-shot & GenTune & 0.821 & -- \\
& & 0-shot & LitTune & -- & 0.826 \\
& & 0-shot & MixTune & 0.796 & 0.808 \\
& & 2-shot & GenTune & \textbf{0.844} & -- \\
& & 2-shot & LitTune & -- & 0.828 \\
& & 2-shot & MixTune & 0.827 & 0.825 \\
\bottomrule
\end{tabularx}
\caption{Main summary results for all 30 tuned SLMs, reported as positive-class micro-F1 (no-relation class excluded), dataset-macro averaged over the seven General benchmarks and the two Literature benchmarks. GenTune and LitTune are evaluated only on their respective domain (the other domain shows ``--''); MixTune is evaluated on both. The single highest General Avg and the single highest Literature Avg are shown in bold. $^\dagger$SmolLM3-3B MixTune 0-shot is reported under the pre-specified default protocol: this reasoning model emits \texttt{<think>} tokens in place of a label and scores \textbf{0} (the value shown). As a post-hoc rescue, disabling reasoning at inference, chat-template flag \texttt{enable\_thinking=False}, plus stripping any residual \texttt{<think>}$\ldots$\texttt{</think>} span, recovers a weak 0.18, and the 2-shot prompt removes the behavior entirely. $^\ddagger$Qwen2.5-3B GenTune 0-shot generates labels from incorrect relation schemas (e.g., Wikidata labels on TACRED), indicating poor schema grounding without few-shot demonstrations.}
\label{tab:main_summary_results}
\end{table*}

Several patterns emerge from the results. First, scale is not the sole determinant of performance: the sub-billion Qwen2.5-0.5B reaches 0.828 General Avg under 2-shot GenTune, matching the same-regime Qwen2.5-3B (0.824) and trailing the 3B models SmolLM3-3B (0.833) and Llama-3.2-3B (0.844) by half a point to under two points. Second, scale and training regime both matter, in complementary ways: the top configurations are 3B models (Llama-3.2-3B leads general RE at 0.844 GenTune 2-shot; SmolLM3-3B leads literary RE at 0.833 LitTune 0-shot), so larger models occupy the top of the table, though this cross-family gap is confounded with model family and is small within the cleanest same-generation contrast (Section~\ref{subsec:scale_prompt}), while within a given size it is the tuning regime that aligns a model to its domain. Third, when the requirement is a single model that handles both domains, MixTune is the strongest choice: Llama-3.2-3B MixTune 2-shot maintains 0.827 on general and 0.825 on literary RE simultaneously, close to each specialist's in-domain peak (see the specialist-versus-generalist analysis below).

A further pattern concerns the interaction between scale and prompt format. For the sub-billion models, 2-shot tuning helps in every regime without exception (Table~\ref{tab:prompt_effect_deltas}), whereas for the 3B models the effect is small and occasionally negative: SmolLM3-3B LitTune and Qwen2.5-3B MixTune both score slightly higher at 0-shot than at 2-shot. A plausible explanation is that larger models absorb the extraction schema more completely from supervised fine-tuning alone, so additional in-prompt demonstrations add little signal and can even misorient the model or consume context that would otherwise support extraction. We return to this asymmetry quantitatively in Section~\ref{subsec:scale_prompt}.

Two configurations are notable 0-shot outliers. Under the pre-specified default protocol, SmolLM3-3B MixTune 0-shot emits \texttt{<think>} reasoning tokens in place of a label and scores \emph{zero}, its primary result; as a post-hoc rescue, disabling reasoning at inference (the chat-template flag \texttt{enable\_thinking=False}, plus stripping any residual \texttt{<think>}$\ldots$\texttt{</think>} span) recovers a valid but weak 0.18 F1, and the 2-shot prompt removes the behavior entirely. Qwen2.5-3B GenTune 0-shot reaches only 0.28 F1, generating labels from incorrect relation schemas. Both are suppressed by 2-shot prompting, highlighting the interaction between model architecture and prompt format. Because both reflect a decoding- or template-level artifact rather than relation-extraction capability, we adopt a pre-specified rule: the two configurations are excluded from the scale-averaged prompt-effect decomposition, the generation-format comparison, and the scaling trend (Table~\ref{tab:shot_decomposition}, Table~\ref{tab:constrained_vs_open_full}, Figure~\ref{fig:scaling_trends}), and are flagged wherever they appear in per-configuration tables (Table~\ref{tab:prompt_effect_deltas}, Appendix~\ref{app:full_results}). Including them in the aggregates would inflate the apparent 0-to-2-shot gains but does not change their direction or statistical significance. For context, frontier LLMs evaluated zero-shot on the same test sets achieve General Avg positive-class F1 of 0.69 (GPT-5.4) and 0.66 (Claude Sonnet~4.6) (Table~\ref{tab:general_frontier_llm_comparison}), so every well-tuned SLM in our grid, including the sub-billion Qwen2.5-0.5B, surpasses the strongest frontier system on general-domain RE, a large proprietary model with an undisclosed parameter count. The per-dataset breakdown below qualifies this comparison, and the full frontier comparison on literary benchmarks is presented in Section~\ref{sec:literary_re}.

\paragraph{General-domain frontier comparison.}
To substantiate the comparison with frontier systems on general-domain RE, Table~\ref{tab:general_frontier_llm_comparison} reports the per-dataset breakdown for the two strongest tuned configurations against GPT-5.4 and Claude Sonnet~4.6 under the same zero-shot schema-enumerated protocol. Averaged over the seven general benchmarks, the best tuned models exceed both frontier systems (Llama-3.2-3B GenTune 2-shot at 0.844 and Qwen2.5-0.5B GenTune 2-shot at 0.828, versus 0.693 for GPT-5.4 and 0.662 for Claude Sonnet~4.6, scored on the full test sets). The advantage is broad: the best tuned SLM leads on all seven datasets. It is largest on schema-heavy benchmarks such as REBEL (0.92 vs.\ 0.68) and Re-DocRED (0.74 vs.\ 0.56), where task-specific supervision matters most, and it is preserved even on the small, knowledge-oriented schemas. In particular GIDS, which an earlier subsampled frontier evaluation had appeared to favour, becomes an SLM win once the frontier models are scored on the full test set (0.85--0.88 for the tuned SLMs vs.\ 0.79 for GPT-5.4). Frontier performance no longer exceeds the tuned SLMs on any general benchmark.

Because the two strongest configurations use two in-context demonstrations at inference whereas the frontier models are evaluated zero-shot, we also report a demonstration-matched comparison. The best 0-shot-tuned model, Llama-3.2-3B GenTune 0-shot, receives no demonstrations, exactly as the frontier models do, yet still reaches 0.821 General Avg and exceeds both GPT-5.4 (0.693) and Claude Sonnet~4.6 (0.662) on all seven datasets (Table~\ref{tab:general_frontier_llm_comparison}). The SLM advantage on general-domain RE therefore stems from task-specific fine-tuning rather than from the in-context demonstrations, whose separate contribution we isolate in Section~\ref{subsec:scale_prompt}.

% General-domain frontier comparison
\begin{table*}[t]
\centering
\small
\setlength{\tabcolsep}{4.5pt}
\renewcommand{\arraystretch}{1.12}
\begin{tabularx}{\textwidth}{l*{8}{>{\centering\arraybackslash}X}}
\toprule
\textbf{Model} & \textbf{TACRED} & \textbf{SemEval} & \textbf{CoNLL04} & \textbf{NYT11} & \textbf{GIDS} & \textbf{Re-DocRED} & \textbf{REBEL} & \textbf{General Avg} \\
\midrule
Llama-3.2-3B GenTune 2s & \textbf{0.694} & \textbf{0.880} & \textbf{0.995} & \textbf{0.828} & 0.844 & \textbf{0.743} & \textbf{0.921} & \textbf{0.844} \\
Qwen2.5-0.5B GenTune 2s & 0.647 & 0.856 & 0.993 & 0.792 & \textbf{0.883} & 0.724 & 0.901 & 0.828 \\
Llama-3.2-3B GenTune 0s & 0.630 & 0.864 & 0.988 & 0.812 & 0.820 & 0.727 & 0.906 & 0.821 \\
\midrule
GPT-5.4 (0-shot) & 0.531 & 0.710 & 0.926 & 0.649 & 0.791 & 0.559 & 0.684 & 0.693 \\
Claude Sonnet 4.6 (0-shot) & 0.445 & 0.709 & 0.919 & 0.692 & 0.799 & 0.523 & 0.547 & 0.662 \\
\bottomrule
\end{tabularx}
\caption{Per-dataset positive-class micro-F1 (no-relation class excluded) on the seven general-domain benchmarks: strongest tuned SLMs versus frontier general-purpose LLMs, all scored on the full test sets. Frontier models are evaluated zero-shot via the OpenRouter API at each provider's \emph{default} reasoning effort (\texttt{none} for GPT-5.4 per its model card; Appendix~\ref{app:frontier_prompts}), with failed or empty generations counted as errors. The best tuned SLM exceeds both frontier systems on the General Avg and on all seven datasets. Best value per column in \textbf{bold}. Gemini~2.5 Pro is omitted because it rarely produced schema-valid output on the general benchmarks (valid-schema rate $<$0.11). Unlike an earlier subsampled evaluation, frontier outputs here are scored on the same full test instances as the SLMs. The Llama-3.2-3B GenTune 0-shot row is a demonstration-matched reference: evaluated with no in-context examples, exactly as the frontier models are, it still exceeds both frontier systems on every dataset.}
\label{tab:general_frontier_llm_comparison}
\end{table*}

\paragraph{Comparison with supervised baselines.}
Although our primary comparison is with frontier LLMs, the positive-class metric also lets us situate the tuned SLMs against fully-supervised systems on the two benchmarks with established evaluation protocols. On TACRED, scored with the conventional micro-F1 that excludes \texttt{no\_relation}, the best tuned SLM reaches 0.71 (SmolLM3-3B GenTune 2-shot), on par with dedicated supervised encoders such as SpanBERT (0.71) and below LUKE (0.73), and well above zero-shot GPT-5.4 (0.53); we note that those encoders use typed entity markers and full supervision, whereas our models are mention-only and trained within a 200k-example cap, and that this comparison is on the same given-entity-pair classification setup that TACRED and SemEval define, so it is task-comparable, unlike the document-level extraction leaderboards we omit below. On SemEval-2010 Task~8, scored with the official direction-aware macro-F1 over the nine relations, the tuned SLMs reach 0.88, on par with strong supervised baselines such as R-BERT~\cite{wu2019enriching} (0.89) and far above the frontier models (0.70). Scored by each benchmark's own convention, then, our compact models are competitive with or close to supervised state of the art while remaining far cheaper than the frontier LLMs that are our main reference point. We report published baseline numbers rather than re-running these systems, and the comparison is therefore approximate (setup and entity-marker conventions differ); for the document-level benchmarks (Re-DocRED, REBEL) and the distantly-supervised~\cite{riedel2010modeling} GIDS, the native evaluation differs from our sentence-level relation-classification setup, so we do not place our numbers against their leaderboards.

\paragraph{Specialist vs.\ generalist supervision.}
Each specialist achieves the highest score in its own domain, since its supervision is aligned to that domain's relation schemas and text style. The question of practical interest is therefore not whether specialization helps in-domain, but how much in-domain performance a single mixed-domain model must give up in exchange for covering both domains at once. We did not evaluate the specialists outside their training domain (Section~\ref{subsec:training}), so we make no claim about how far a specialist degrades under domain shift; the comparison we can make rigorously is between each specialist's in-domain peak and MixTune's simultaneous performance on the same domain.

The specialist regimes achieve the highest in-domain scores: GenTune reaches 0.772 averaged over the seven general-domain benchmarks, and LitTune reaches 0.789 over the two literary benchmarks. These peaks reflect supervision that is aligned to each domain's relation schemas (e.g., TACRED's 41 relation types and Re-DocRED's 96 for general RE; Biographical's \texttt{educatedAt} and PG-Fiction's fine-grained ontology for literary RE) and to its characteristic text style, expository in the general case and narrative in the literary one.

MixTune trades a modest amount of this in-domain performance for the ability to cover both domains with a single model. It reaches 0.753 on general and 0.763 on literary RE (within two to three points of the respective specialist averages) with a domain-balance gap of just 0.010 between the two domains. In other words, one mixed-domain model comes close to matching each specialist on its own ground without the need to select, store, or serve a separate model per domain. If the deployment objective is a single small model capable of relation extraction across varied domains, MixTune is the most attractive option: the cost is a few points of in-domain peak performance, and the benefit is balanced coverage that a single specialist cannot provide. We make this argument from the per-domain scores directly, without a pooled ``overall'' average, because pooling would average over a different mix of in-domain and out-of-domain datasets for each regime and would not be comparable across regimes.

\begin{figure*}[!tbp]
 \centering
 \includegraphics[width=\textwidth]{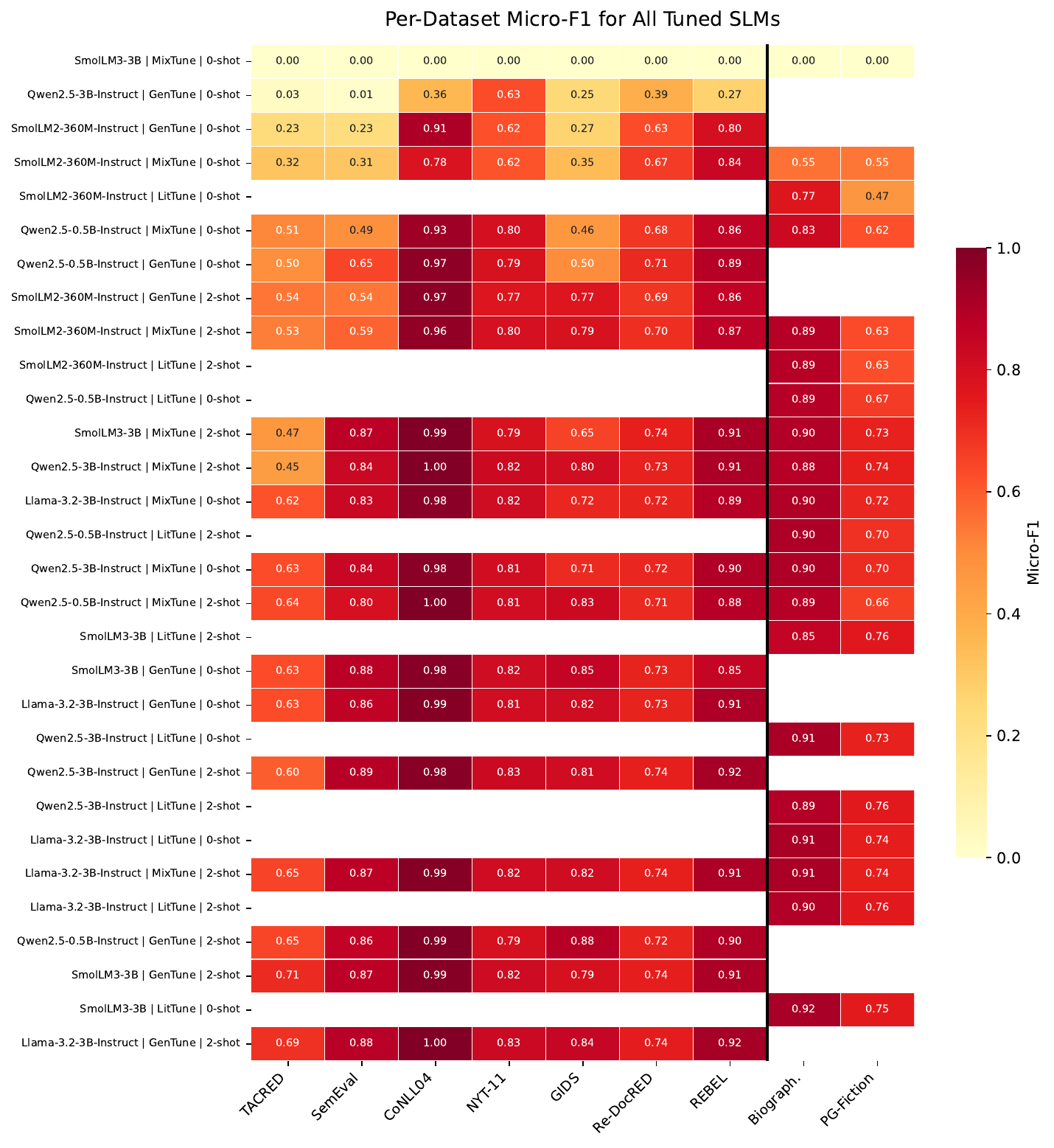}
 \caption{Heatmap of per-dataset positive-class micro-F1 for all 30 tuned SLMs across the nine evaluation benchmarks, under schema-enumerated prompting with matched prompt shots (consistent with Tables~\ref{tab:main_summary_results} and~\ref{tab:full_results_matrix}). Rows correspond to tuned model configurations and columns to datasets, separated into general and literary groups. The figure highlights domain specialization and the relative balance of mixed-domain tuning across benchmarks.}
 \label{fig:heatmap_all_models}
\end{figure*}

\paragraph{Dataset-level interpretation.}
Figure~\ref{fig:heatmap_all_models} provides a per-dataset heatmap that reveals where the domain averages mask important variation. Among the general-domain datasets, CoNLL04 and REBEL are near-saturated (positive-class F1 $>$ 0.92 for most 3B models), whereas TACRED is the hardest (mean positive-class F1 $\sim$0.58), followed by SemEval, GIDS, and Re-DocRED (0.72--0.75), suggesting that schema ambiguity and label granularity matter more than domain alone. On the literary side, Biographical yields consistently higher scores than PG-Fiction across all tuning regimes, reflecting Biographical's smaller ontology (10 relations vs.\ 137) and more formulaic sentence structure. The per-dataset view confirms that MixTune's balanced both-domain performance is broadly distributed across benchmarks rather than driven by any single one.

In short, scale and regime contribute jointly to performance, domain specialists attain the highest in-domain averages (GenTune 0.772 general, LitTune 0.789 literary), and, most useful for deployment, a single MixTune model covers both domains at a domain-balance gap of only 0.010, sacrificing about two points relative to each specialist.

% ------------------------------------------------------------------------------
\subsection{Effects of Scale and Prompt-Conditioned Supervision}
\label{subsec:scale_prompt}
We next examine three potential sources of the performance gains observed above: model scale, prompt-conditioned supervision, and their interaction. If gains are explained only by increasing parameter count, the practical value of the training strategy is limited; if prompt-conditioned supervision meaningfully improves smaller models, the results provide stronger evidence for efficient and democratized RE.

\begin{figure*}[!tbp]
 \centering
 \includegraphics[width=\textwidth]{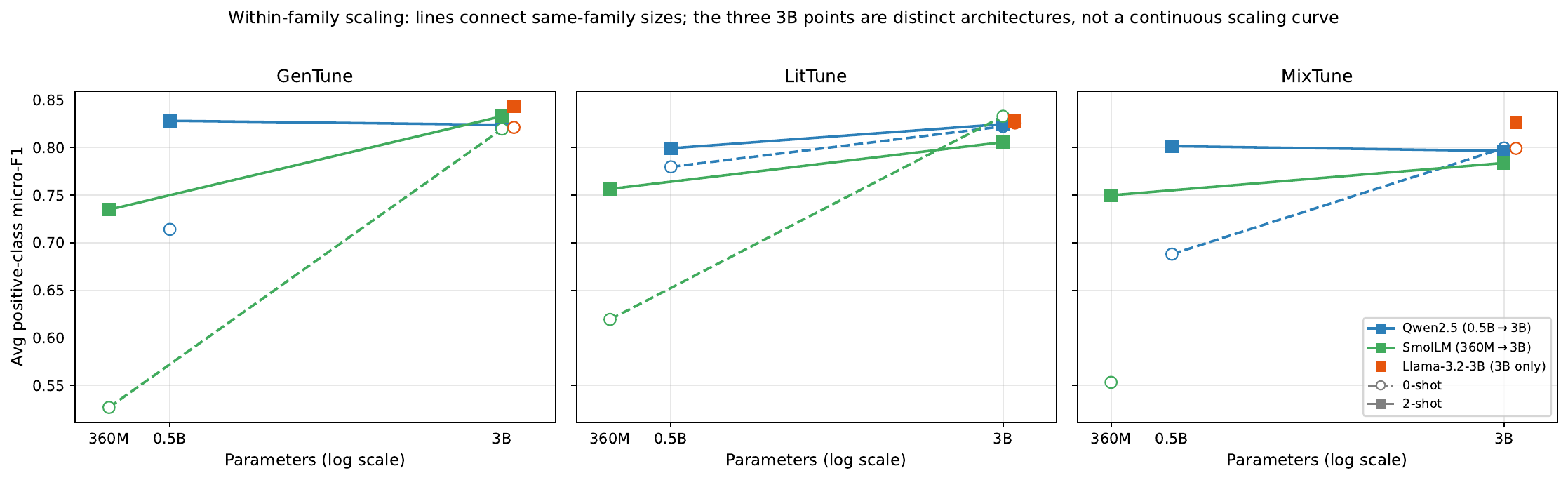}
 \caption{Within-family effect of scale on average positive-class micro-F1 (over each regime's evaluation datasets), under schema-enumerated prompting with matched prompt shots. Because parameter count is confounded with model family, lines connect \emph{only} same-family sizes, Qwen2.5 ($0.5$B$\to$$3$B) and SmolLM (SmolLM2-360M$\to$SmolLM3-3B), and Llama-3.2-3B is shown as an isolated $3$B point (it has no sub-billion counterpart); we deliberately do not draw a trend across the three distinct $3$B architectures. Dashed lines with open markers are 0-shot; solid lines with filled markers are 2-shot. The two 0-shot anomalies (SmolLM3-3B MixTune, Qwen2.5-3B GenTune; Table~\ref{tab:main_summary_results}) are excluded, leaving a single endpoint where applicable. Same-generation Qwen2.5 scaling is small (flat on GenTune), whereas the larger SmolLM gain also reflects a SmolLM2$\to$SmolLM3 generation change (Table~\ref{tab:within_family_scale}).}
 \label{fig:scaling_trends}
\end{figure*}

\paragraph{Scaling effects (within family).}
Parameter count in our grid is confounded with model family: the five base models differ in tokenizer, pretraining corpus, and instruction tuning, and we have only two broad scales (sub-billion and 3B) with no sub-billion Llama. We therefore read scale only \emph{within} a family, where size varies while the rest of the architecture is held roughly fixed, and avoid causal claims about ``capacity.'' Two within-family contrasts are available (Table~\ref{tab:within_family_scale}): Qwen2.5 ($0.5$B$\to$$3$B, the same generation) and SmolLM (SmolLM2-360M$\to$SmolLM3-3B, which additionally crosses a model generation); Llama-3.2-3B has no sub-billion counterpart and cannot inform a within-family slope. The two contrasts are heterogeneous. Scaling Qwen2.5, the cleanest, same-generation contrast, adds only $+0.037$ overall positive-class micro-F1 (95\% CI $[+0.009,+0.067]$) and \emph{nothing} on general-domain GenTune ($-0.004$), whereas scaling SmolLM adds more ($+0.132$) but conflates size with a generation of improved pretraining and data. A regression of micro-F1 on $\log_{10}$(parameters) gives a slope of $+0.129$ per $10\times$ parameters $[+0.077,+0.196]$; adding family fixed effects barely changes it ($+0.114$ $[+0.065,+0.180]$), but this pooled slope assumes a common within-family scale effect that the data do not support (the two families' within-family effects differ by more than $3\times$, $+0.037$ vs.\ $+0.132$), so it is driven largely by the generation-confounded SmolLM contrast. We therefore describe larger size as \emph{associated} with higher F1 within a family, clearly for SmolLM, only weakly for same-generation Qwen2.5, rather than as a continuous law across these unrelated architectures, and Figure~\ref{fig:scaling_trends} accordingly connects only same-family sizes. The practical reading is favorable: even a $6\times$ scale-up within Qwen2.5 buys little on general-domain RE, so sub-billion models remain strong deployment options.

\begin{table}[t]
\centering
\small
\setlength{\tabcolsep}{4pt}
\renewcommand{\arraystretch}{1.12}
\begin{tabular}{llcccc}
\toprule
\textbf{Family} & \textbf{Scale} & \textbf{Gen} & \textbf{Lit} & \textbf{Mix} & \textbf{Overall [95\% CI]} \\
\midrule
Qwen2.5 & 0.5B$\to$3B & $-0.004$ & $+0.034$ & $+0.053$ & $+0.037$ $[+0.009,+0.067]$ \\
SmolLM$^{\ast}$ & 360M$\to$3B & $+0.195$ & $+0.131$ & $+0.034$ & $+0.132$ $[+0.070,+0.215]$ \\
\bottomrule
\end{tabular}
\caption{Within-family effect of scale on positive-class micro-F1 ($\Delta = $ 3B $-$ sub-billion), paired by (regime, shot, dataset) on the primary schema-enumerated, matched-shot subset with the two 0-shot anomalies excluded; Gen/Lit/Mix are the per-regime means and CIs are dataset-clustered bootstrap (10k resamples). Llama-3.2-3B is omitted (no sub-billion counterpart). $^{\ast}$The SmolLM contrast crosses SmolLM2$\to$SmolLM3, so it conflates scale with a generation change in pretraining and data; the same-generation Qwen2.5 contrast is the cleaner scale estimate. A family-controlled regression gives a $\log_{10}$(params) slope of $+0.114$ $[+0.065,+0.180]$ per $10\times$ parameters (naive cross-family slope $+0.129$ $[+0.077,+0.196]$). Computed by \texttt{scripts/analyze\_scale\_family.py}.}
\label{tab:within_family_scale}
\end{table}

The practical implication is clear. When the performance gap between 0.5B and 3B models is small, as in GenTune, sub-billion models become realistic deployment options. For example, Qwen2.5-0.5B occupies under 1\,GB in BF16 (roughly 0.5\,GB as a 4-bit backbone, or 0.3\,GB as a \texttt{Q4\_K\_M} GGUF), small enough for CPU deployment and, plausibly, for mobile NPUs, though we do not benchmark NPU inference, and still achieves 0.83 positive-class F1 on general-domain RE.
However, the gap between the best 3B and best sub-billion configuration is larger on a few benchmarks, notably GIDS and SemEval, which require broader world knowledge and finer-grained relation disambiguation. Because that best-in-class gap again spans different families, we read it as a cross-family difference rather than a pure scale effect, but it marks where training strategy alone does not close the distance between the strongest small and 3B models.

\paragraph{Prompt-conditioned tuning effects.}
Table~\ref{tab:prompt_effect_deltas} reports the F1 delta between each 2-shot-tuned model, evaluated with a 2-shot prompt, and its 0-shot-tuned counterpart, evaluated with a 0-shot prompt. Because the tuning shot and the prompt shot are matched on both sides, this delta measures the combined effect of the two-demonstration pipeline, demonstrations present both during fine-tuning and at inference, relative to the zero-demonstration pipeline; it does not by itself isolate the training-time contribution, which we separate in the prompt-shot decomposition below.

% Table 5
% Table 5. Effect of prompt-conditioned tuning: 2-shot minus 0-shot
\begin{table*}[t]
\centering
\small
\setlength{\tabcolsep}{5pt}
\renewcommand{\arraystretch}{1.12}
\begin{tabularx}{\textwidth}{ll*{3}{>{\centering\arraybackslash}X}}
\toprule
\textbf{Base Model} & \textbf{Tuning Regime} & \textbf{$\Delta$ General Avg F1} & \textbf{$\Delta$ Literature Avg F1} & \textbf{$\Delta$ Avg F1} \\
\midrule

\multirow{3}{*}{SmolLM2-360M-Instruct}
& GenTune & \textbf{+0.208} & -- & +0.208 \\
& LitTune & -- & +0.137 & +0.137 \\
& MixTune & +0.192 & \textbf{+0.212} & \textbf{+0.197} \\
\midrule

\multirow{3}{*}{Qwen2.5-0.5B-Instruct}
& GenTune & +0.114 & -- & +0.114 \\
& LitTune & -- & +0.020 & +0.020 \\
& MixTune & +0.132 & +0.048 & +0.113 \\
\midrule

\multirow{3}{*}{SmolLM3-3B}
& GenTune & +0.013 & -- & +0.013 \\
& LitTune & -- & $-$0.027 & $-$0.027 \\
& MixTune & +0.775$^\dagger$ & +0.816$^\dagger$ & +0.784$^\dagger$ \\
\midrule

\multirow{3}{*}{Qwen2.5-3B-Instruct}
& GenTune & +0.547$^\ddagger$ & -- & +0.547$^\ddagger$ \\
& LitTune & -- & +0.002 & +0.002 \\
& MixTune & $-$0.006 & +0.006 & $-$0.003 \\
\midrule

\multirow{3}{*}{Llama-3.2-3B-Instruct}
& GenTune & +0.022 & -- & +0.022 \\
& LitTune & -- & +0.002 & +0.002 \\
& MixTune & +0.030 & +0.016 & +0.027 \\
\bottomrule
\end{tabularx}
\caption{Performance gain in positive-class micro-F1 of the matched two-demonstration pipeline (2-shot tuning evaluated with a 2-shot prompt) relative to the matched zero-demonstration pipeline (0-shot tuning evaluated with a 0-shot prompt). Positive values therefore reflect the \emph{joint} effect of demonstrations during fine-tuning and at inference, not the training-time contribution in isolation; Section~\ref{subsec:scale_prompt} decomposes the two. $^\dagger$Computed against SmolLM3-3B's default-protocol 0-shot MixTune baseline (0; the \texttt{<think>}-emission artifact of Section~\ref{sec:results}), so this delta reflects the decoding artifact rather than a genuine prompt effect; the configuration is excluded from the scale-averaged decomposition (Table~\ref{tab:shot_decomposition}). $^\ddagger$Inflated by Qwen2.5-3B's schema-confused 0-shot GenTune baseline (Section~\ref{sec:results}); likewise excluded from the scale-averaged decomposition. $\Delta$ Avg F1 is computed over each regime's evaluation datasets (general-only for GenTune, literary-only for LitTune, all nine for MixTune).}
\label{tab:prompt_effect_deltas}
\end{table*}

The results confirm that prompt context is most valuable for smaller models. SmolLM2-360M gains 14--21 F1 points from 2-shot tuning across all three regimes, while the three 3B models gain less than 2 points on average (excluding the anomalous SmolLM3-3B and Qwen2.5-3B 0-shot failures). This asymmetry suggests that demonstrations compensate for limited internal task abstraction at small scale, but become redundant once the model can infer the extraction schema from repeated supervised exposure alone.

\paragraph{Disentangling training-time from inference-time demonstrations.}
The matched delta cannot, on its own, credit the gain to demonstrations seen \emph{during fine-tuning}, since it varies demonstrations at training and inference simultaneously. Evaluating every 2-shot-tuned checkpoint with a 0-shot prompt separates them: writing $F_{t,p}$ for the score at tuning shot $t$ and prompt shot $p$, the matched delta splits additively,
\[
\underbrace{F_{2,2}-F_{0,0}}_{\text{matched}}
=\underbrace{(F_{2,0}-F_{0,0})}_{\text{training-time}}
+\underbrace{(F_{2,2}-F_{2,0})}_{\text{inference-time}}.
\]
Averaged by scale (Table~\ref{tab:shot_decomposition}, non-anomalous configurations), the training-time term is \emph{negative} at both scales ($-0.27$ sub-billion, $-0.14$ at 3B) while the entire positive gain comes from the inference-time term ($+0.41$, $+0.14$): a 2-shot-tuned checkpoint evaluated \emph{without} demonstrations is consistently worse than its 0-shot-tuned counterpart (e.g.\ SmolLM2-360M GenTune $0.527{\to}0.240$). So 2-shot tuning chiefly makes a model \emph{dependent} on inference demonstrations rather than teaching the task, most strongly at sub-billion scale. The training-time term also absorbs a train/inference prompt-format mismatch, so it upper-bounds any intrinsic harm from demonstration-conditioned tuning.

\begin{table}[t]
\centering
\small
\setlength{\tabcolsep}{4pt}
\renewcommand{\arraystretch}{1.12}
\begin{tabularx}{\columnwidth}{l*{6}{>{\centering\arraybackslash}X}}
\toprule
\textbf{Scale} & $F_{0,0}$ & $F_{2,0}$ & $F_{2,2}$ & \textbf{Train} & \textbf{Infer} & \textbf{Matched} \\
\midrule
Sub-billion & 0.647 & 0.373 & 0.778 & $-0.274$ & $+0.406$ & $+0.132$ \\
3B & 0.817 & 0.679 & 0.823 & $-0.138$ & $+0.143$ & $+0.005$ \\
\bottomrule
\end{tabularx}
\caption{Decomposition of the matched 2-shot$-$0-shot gain into a training-time and an inference-time component, in positive-class micro-F1 averaged by scale over the non-anomalous configurations. $F_{t,p}$ is the score of a checkpoint with tuning shot $t$ evaluated at prompt shot $p$; Train $=F_{2,0}-F_{0,0}$ (vary tuning, fix 0-shot prompt), Infer $=F_{2,2}-F_{2,0}$ (fix the 2-shot-tuned checkpoint, vary the prompt), and Matched $=$ Train $+$ Infer. The training-time term is negative at both scales: the matched gain is driven entirely by inference-time demonstrations.}
\label{tab:shot_decomposition}
\end{table}

\paragraph{Interaction between scale and prompt context.}
Considering scale and prompt context jointly reveals that 2-shot tuning partially compensates for limited capacity. SmolLM2-360M with 2-shot MixTune (0.750 averaged across all nine datasets) closes about four-fifths of the gap between its own 0-shot counterpart (0.553) and Llama-3.2-3B with 0-shot MixTune (0.799, same averaging), suggesting that prompt-conditioned supervision is most valuable precisely where it is cheapest, namely on the smallest models that benefit most from explicit task structure. At 3B scale, the interaction reverses: gains from demonstrations are negligible for well-behaved architectures, while the dominant factor becomes data composition (MixTune vs.\ specialist). These findings separate three improvement sources that might otherwise be conflated: scale provides a consistent advantage; prompt-conditioned supervision partially compensates for limited capacity; and data composition determines the ceiling that neither scale nor prompt format can overcome alone.

\paragraph{Truncation does not confound the shot comparison.}
Because 2-shot prompts are longer than 0-shot prompts, one might worry that the 0-shot-versus-2-shot comparison is confounded by truncated context. It is not: all reported scores come from inference, where truncation never fires, the longest input across all datasets, conditions, and tokenizers is $2{,}500$ tokens, far below the smallest $8{,}192$-token context, so no evaluated prompt loses any context. The only truncation anywhere in the pipeline affects at most ${\sim}1.1\%$ of one literary dataset's (PG-Fiction) 2-shot \emph{training} sequences and $0\%$ of all general-domain data; because it is right-side it would, if anything, slightly handicap the longer 2-shot condition by dropping its trailing query and gold label, making the comparison conservative against, not inflationary of, the reported 2-shot benefit (Appendix~\ref{app:truncation}).

\paragraph{Schema-enumerated vs.\ generic prompting.}
A natural question is whether injecting the allowed label set into the system prompt (schema-enumerated prompting) improves extraction quality relative to a generic prompt that does not enumerate labels (generic prompting). Because all models were fine-tuned on RE-specific data with well-defined ontologies, they may already internalize schema-faithful generation during training, making the additional label-set prompt redundant or even distracting. We test this by comparing the same model checkpoints under both prompting regimes.

\begin{figure}[htbp]
 \centering
 \includegraphics[width=\columnwidth]{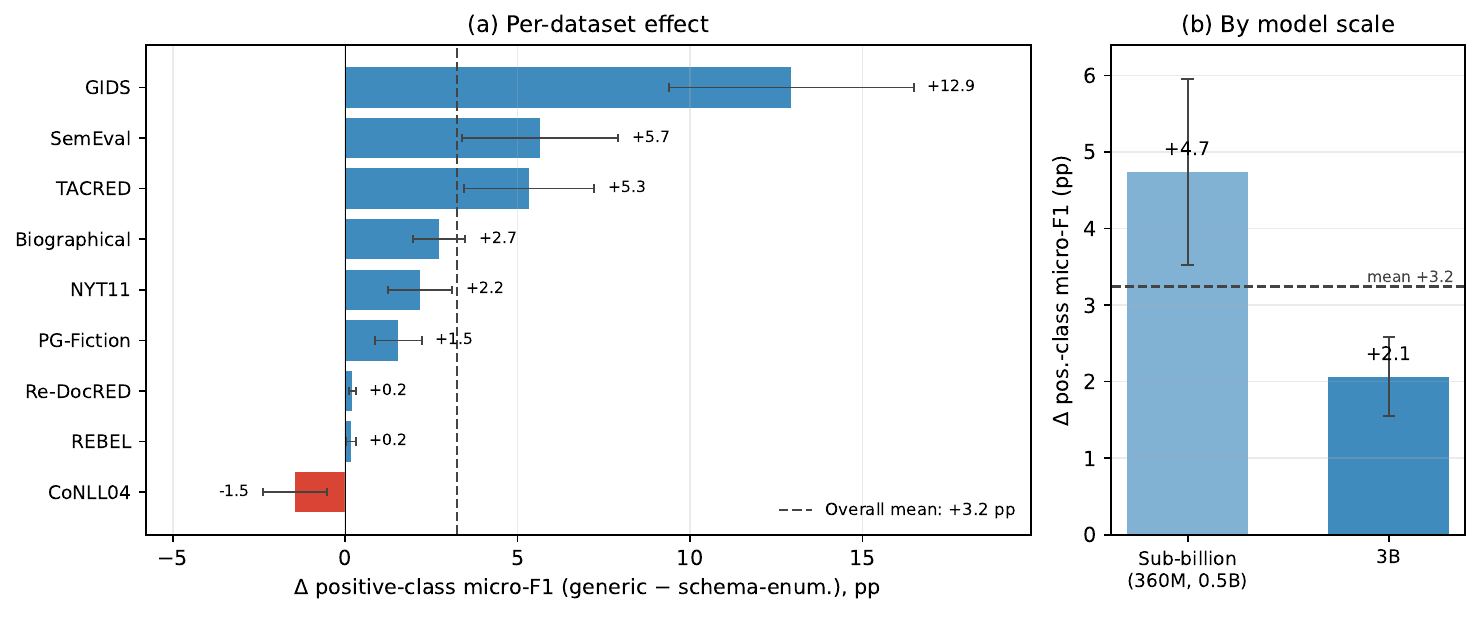}
 \caption{Effect of prompting condition on \textbf{positive-class micro-F1} (no-relation class excluded), matched-shot evaluations only. (a)~Per-dataset $\Delta$ (generic $-$ schema-enumerated); positive values favor generic prompting. Error bars show the standard error of the mean; the dashed line marks the overall mean ($+3.2$~pp). (b)~Breakdown by model scale: sub-billion models benefit more ($+4.7$~pp) than 3B models ($+2.1$~pp). The all-class accuracy gain is comparable ($+3.1$~pp), so the improvement is not driven by the majority negative class.}
 \label{fig:constrained_vs_open}
\end{figure}

Contrary to expectation, generic prompting outperforms schema-enumerated prompting on the primary positive-class micro-F1 metric by an average of $+3.2$ points across the 164 paired matched-shot evaluations (excluding the two known anomalous configurations), and the gain is larger for sub-billion models ($+4.7$) than for 3B models ($+2.1$). Crucially, this is \emph{not} an artifact of the majority no-relation class: the all-class accuracy gain ($+3.1$ points) is essentially identical, and on the most negative-heavy benchmarks (TACRED $+5.3$, Biographical $+2.7$) the positive-class gain is as large as or larger than the accuracy gain, so the effect reflects better positive-relation extraction rather than improved no-relation prediction. Generic prompting helps on 8 of the 9 datasets; the sole exception is CoNLL04 ($-1.5$, a 5-label schema), and the largest gain is on GIDS ($+12.9$), whose tiny 4-relation schema combined with high label ambiguity appears to benefit from the model's internalized schema rather than runtime enumeration. Output well-formedness is essentially unaffected: schema-valid rates are 0.882 (schema-enumerated) versus 0.875 (generic), and malformed-output rates are negligible under both ($<$0.1\%), so label-set enumeration provides no formatting benefit that offsets its F1 cost. Figure~\ref{fig:constrained_vs_open} visualizes the per-dataset and by-scale effect; full per-dataset scores are reported in Appendix~\ref{app:constrained_vs_open}.

This finding has two implications. First, it validates the quality of the fine-tuning procedure: supervised RE training with well-structured prompts is sufficient to teach schema compliance without runtime label enumeration. Second, it motivates a practical deployment recommendation: for fine-tuned SLMs, generic prompting yields higher positive-class F1, lower latency (shorter prompts), and no loss of output well-formedness, making it the preferable mode in deployment. We retain schema-enumerated prompting as the primary evaluation protocol throughout the paper because it is the more conservative and reproducible setting; because generic prompting raises positive-class micro-F1 by $+3.2$ on average (on 8 of 9 datasets), the reported schema-enumerated scores are conservative lower bounds on what the same checkpoints achieve.

% ------------------------------------------------------------------------------
\subsection{Literary Relation Extraction}
\label{sec:literary_re}
Literary RE remains the most challenging setting in our evaluation. PG-Fiction yields the lowest scores of any benchmark (at most 0.76 positive-class micro-F1, versus 0.99 on CoNLL04 and 0.92 on REBEL), because relation expression in narrative text is often implicit and discourse-dependent. Under the full 137-label inventory, performance on the long tail is far weaker than the micro-averages suggest (positive-class macro-F1 is only ${\sim}$0.42 for the best 3B models). Much of this shortfall, however, stems from a long tail of out-of-schema annotations: 90 of the 137 labels, just 2.3\% of positive instances, fall outside the dataset's documented 48-relation ontology, and re-scoring against that canonical ontology leaves micro-F1 essentially unchanged (${+}0.01$) while raising macro-F1 to ${\sim}0.68$ (Table~\ref{tab:pgfiction_dual}; analyzed in the dual-ontology paragraph below). We therefore include two targeted follow-up analyses: a DAPT extension that tests whether continued adaptation to literary text improves performance beyond supervised fine-tuning alone, and a frontier LLM comparison that contextualizes the remaining headroom.

\paragraph{Targeted DAPT follow-up.}
We first examine whether continued adaptation to literary text yields benefits beyond supervised literature-domain fine-tuning alone. Although DAPT is not part of the main 30-model experimental grid, it is a plausible extension for literary RE because narrative corpora differ substantially from the expository and benchmark-style text on which most instruction-tuned models are optimized. To test this hypothesis we apply \emph{full-parameter} domain-adaptive pretraining, continued causal-language-model training on the full model weights, not an adapter-based or quantized procedure, on LitBank ($\sim$80M tokens; Table~\ref{tab:hyper_dapt}) to Llama-3.2-3B, the strongest 3B-class model in our grid, and then fine-tune the resulting checkpoint with QLoRA under two regimes: LitTune (literary labels only) and MixTune (balanced general + literary labels). This yields a clean 2$\times$2 design (Table~\ref{tab:dapttune_literature}) that crosses (DAPT vs.\ no-DAPT) with (LitTune vs.\ MixTune) on the Biographical and PG-Fiction benchmarks. We omit GenTune from this comparison because a model adapted to literary discourse and then fine-tuned exclusively on general-domain labels would conflate two opposing signals, making the DAPT contribution uninterpretable. The LitTune pair provides the cleanest test of whether unsupervised literary exposure adds value on top of literary supervision, while the MixTune pair tests whether any DAPT benefit persists when the model also receives general-domain training signal.

% Table 6
\begin{table*}[t]
\centering
\small
\setlength{\tabcolsep}{5pt}
\renewcommand{\arraystretch}{1.12}
\begin{tabularx}{\textwidth}{l l >{\centering\arraybackslash}X >{\centering\arraybackslash}X >{\centering\arraybackslash}X >{\centering\arraybackslash}X}
\toprule
\textbf{Model} & \textbf{Prompt Style} & \textbf{Biographical F1} & \textbf{PG-Fiction F1} & \textbf{Literature Avg F1} & \textbf{$\Delta$ Lit.\ Avg F1} \\
\midrule
Llama-3.2-3B LitTune 0s & 0-shot & 0.912 & 0.740 & 0.826 & -- \\
Llama-3.2-3B-lit-dapt LitTune 0s & 0-shot & 0.912 & 0.742 & 0.827 & +0.001 \\
\midrule
Llama-3.2-3B MixTune 0s & 0-shot & 0.900 & 0.716 & 0.808 & -- \\
Llama-3.2-3B-lit-dapt MixTune 0s & 0-shot & 0.901 & 0.716 & 0.809 & $+$0.001 \\
\bottomrule
\end{tabularx}
\caption{DAPT ablation: effect of literary domain-adaptive pretraining on Llama-3.2-3B under two tuning regimes, reported in positive-class micro-F1 under the identical evaluation protocol of Section~\ref{subsec:evaluation}. The 2$\times$2 design crosses (DAPT vs.\ no-DAPT) with (LitTune vs.\ MixTune), isolating the marginal value of continued literary adaptation beyond supervised fine-tuning. The effect is small under both regimes: at most 0.002 on any individual benchmark (PG-Fiction under LitTune, $0.740{\rightarrow}0.742$) and at most 0.001 in Literature-average F1.}
\label{tab:dapttune_literature}
\end{table*}

Under this single LitBank DAPT configuration, continued literary pretraining provides no practically meaningful benefit. On Biographical, where non-DAPT Llama already achieves $>$0.91 positive-class F1, the change is negligible ($|\Delta| < 0.001$ under both regimes), consistent with near-saturation on this dataset's constrained ontology. On PG-Fiction, the DAPT variant matches its non-DAPT counterpart to within 0.2 percentage points under both regimes, and a per-class analysis over all 137 positive relation types shows essentially unchanged behavior (macro-F1 0.351~$\rightarrow$~0.356; support-weighted F1 0.730~$\rightarrow$~0.735), with the largest per-class movements confined to relations with fewer than ten test instances. Both models also handle the catch-all class identically, abstaining on all 1{,}805 \textit{none}-class examples. We therefore find that, in this setting, supervised fine-tuning on literary RE data already captures much of the relevant domain signal, and roughly 80M tokens of additional unsupervised exposure to literary text adds no practically meaningful gain on top of it. This negative result is informative in two ways. First, it suggests that, for this model, the large advantage over frontier models (see the frontier comparison below) is driven by supervised task adaptation rather than by this continued-pretraining step. Second, we caution against reading it as evidence of \emph{no} corpus overlap; in fact, overlap exists. A near-duplicate analysis, matching verbatim word spans, since PG-Fiction carries no document identifiers, finds that roughly 9\% of PG-Fiction test passages share verbatim text with the LitBank corpus (robust across 10- and 15-word spans), both being drawn from the same canonical Project Gutenberg novels (e.g., \emph{Persuasion}, \emph{Moby Dick}, \emph{Tess of the d'Urbervilles}). That continued pretraining yields no gain even though the DAPT model trains on the full text of books that supply part of the evaluation suggests the base model had already internalized these canonical works during its original pretraining, rather than that overlap is absent. We treat this as a limitation of the PG-Fiction benchmark (Section~\ref{sec:conclusion}).

\paragraph{Frontier-LLM comparison.}
We next compare the strongest SLM configurations against frontier proprietary LLMs on literary RE. This comparison establishes a reference for what large general-purpose systems achieve without task-specific fine-tuning, contextualizing the competitiveness of our tuned 3B models against large proprietary systems with undisclosed parameter counts. The exact prompts and generation parameters used for frontier evaluation are detailed in Appendix~\ref{app:frontier_prompts}.

% Table 7
\begin{table*}[t]
\centering
\small
\setlength{\tabcolsep}{5pt}
\renewcommand{\arraystretch}{1.12}
\begin{tabularx}{\textwidth}{l*{5}{>{\centering\arraybackslash}X}}
\toprule
\textbf{Model} & \textbf{Biographical F1} & \textbf{PG-Fiction F1} & \textbf{Lit.\ Avg F1} & \textbf{Lit.\ Macro-F1} & \textbf{Valid-schema rate} \\
\midrule
SmolLM3-3B LitTune 0s & \textbf{0.917} & \textbf{0.749} & \textbf{0.833} & \textbf{0.65} & \textbf{0.999} \\
Llama-3.2-3B MixTune 2s & 0.913 & 0.737 & 0.825 & 0.62 & 0.999 \\
\midrule
GPT-5.4 (0-shot) & 0.832 & 0.324 & 0.578 & 0.50 & 0.867 \\
Claude Sonnet 4.6 (0-shot) & 0.725 & 0.334 & 0.530 & 0.41 & 0.903 \\
\bottomrule
\end{tabularx}
\caption{Positive-class micro-F1 (no-relation class excluded) for the strongest tuned SLMs versus frontier general-purpose LLMs on the literature benchmarks, all scored on the full test sets. Frontier models are evaluated zero-shot via the OpenRouter API at each provider's \emph{default} reasoning effort (\texttt{none} for GPT-5.4), with failed or empty generations counted as errors (Appendix~\ref{app:frontier_prompts}). The best tuned SLMs substantially outperform both frontier models, which are large proprietary systems with undisclosed parameter counts. The gap persists under positive-class macro-F1 (the \textbf{Lit.\ Macro-F1} column, averaged over the two literary benchmarks), so it is not an artifact of frequent-class dominance. PG-Fiction here uses the full 137-label inventory; Table~\ref{tab:pgfiction_dual} additionally reports the canonical 48-relation ontology, under which the macro-F1 gap over frontier models widens further. Best value per column in bold.}
\label{tab:literature_frontier_llm_comparison}
\end{table*}

The results are striking: tuned SLMs substantially outperform both frontier models on literary RE. SmolLM3-3B LitTune 0-shot achieves a Literature Avg positive-class F1 of 0.833, exceeding GPT-5.4 (0.578) by about 26 points and Claude Sonnet~4.6 (0.530) by about 30 points. Llama-3.2-3B MixTune 2-shot achieves 0.83, similarly dominating both frontier systems. The gap is especially pronounced on PG-Fiction, where the best SLMs reach 0.75 F1 compared to 0.32 (GPT-5.4) and 0.33 (Claude Sonnet~4.6), and 0.76 versus 0.44 and 0.45 under the dataset's canonical 48-relation ontology (Table~\ref{tab:pgfiction_dual}). On Biographical, the margin is narrower but still substantial: 0.92 vs.\ 0.83 and 0.73 respectively. The advantage persists under positive-class macro-F1 (0.65 and 0.62 for the tuned SLMs versus 0.50 and 0.41 for the frontier models), so it is not an artifact of frequent-class dominance. Nor is it an artifact of train/test contamination: although roughly 23\% of Biographical test examples recur verbatim in its training split (the corpus is split per example rather than per document), re-scoring on the de-leaked test set lowers all models comparably, the best tuned SLM from 0.917 to 0.903 and GPT-5.4 from 0.832 to 0.820 on Biographical, so the SLM's margin over the frontier is essentially unchanged; PG-Fiction exhibits no such answer leakage (0.1\%). Schema-valid output rates are high for the tuned SLMs ($>$0.99) but lower for the frontier models on literary text (GPT-5.4 0.87, Claude 0.90; Table~\ref{tab:literature_frontier_llm_comparison}), so part of the frontier deficit reflects malformed or schema-incompatible output rather than relational reasoning alone, a confound that a constrained or structured frontier protocol would remove (Section~\ref{sec:conclusion}).

These findings demonstrate that literary RE is not primarily a scale problem: task-specific fine-tuning on domain-relevant data is substantially more effective than raw model scale, and the DAPT follow-up shows that this supervised adaptation, rather than additional unsupervised exposure to literary text, is what closes the gap. Tuned 3B SLMs surpass large proprietary systems with undisclosed parameter counts by more than 25 F1 points on literary benchmarks.

% Table 8b: PG-Fiction dual-ontology
\begin{table}[t]
\centering
\small
\setlength{\tabcolsep}{6pt}
\renewcommand{\arraystretch}{1.12}
\begin{tabular}{l cc cc}
\toprule
& \multicolumn{2}{c}{\textbf{137-label corpus}} & \multicolumn{2}{c}{\textbf{Canonical 48-relation}} \\
\cmidrule(lr){2-3}\cmidrule(lr){4-5}
\textbf{Model} & micro & macro & micro & macro \\
\midrule
SmolLM3-3B LitTune 0s & 0.749 & 0.415 & 0.757 & 0.680 \\
SmolLM3-3B LitTune 2s & \textbf{0.760} & \textbf{0.438} & \textbf{0.768} & \textbf{0.712} \\
Llama-3.2-3B MixTune 2s & 0.737 & 0.364 & 0.745 & 0.646 \\
\midrule
GPT-5.4 (0-shot) & 0.324 & 0.280 & 0.440 & 0.348 \\
Claude Sonnet 4.6 (0-shot) & 0.334 & 0.231 & 0.448 & 0.373 \\
\bottomrule
\end{tabular}
\caption{PG-Fiction scored under both label inventories: the full 137-label processed corpus and the canonical 48-relation ARF ontology of \citet{christou2024arf}, with the 90 out-of-ontology labels (2.3\% of positive instances) mapped to the background class, exactly as the no-relation class is treated (Appendix~\ref{app:pgfiction48}). Restricting to the canonical ontology leaves micro-F1 essentially unchanged (mean $+0.01$ across all literary configurations) but raises macro-F1 by ${\sim}0.29$, because the 137-label macro-average is dominated by rare out-of-ontology relations. The tuned SLMs' advantage over frontier models \emph{widens} under the canonical ontology on macro-F1 (0.68--0.71 vs.\ 0.35--0.37). Best value per column in \textbf{bold}; full per-configuration scores in Table~\ref{tab:pgfiction48_full}.}
\label{tab:pgfiction_dual}
\end{table}

\paragraph{Canonical-ontology evaluation.}
The low 137-label macro-F1 is largely an artifact of the label inventory. Of the 137 labels, 90 are out-of-ontology relations the GPT-4o annotator emitted despite the fixed 48-relation schema, accounting for just 2.3\% of positive instances; mapping them to the background class (exactly as the no-relation class) yields the canonical 48-relation evaluation in Table~\ref{tab:pgfiction_dual}. Micro-F1 is essentially invariant to this choice (mean ${+}0.01$; Appendix~\ref{app:pgfiction48}), so the ranking does not depend on the inventory, whereas macro-F1 rises sharply ($0.42{\rightarrow}0.68$ for SmolLM3-3B LitTune) once the rare out-of-schema tail is removed. On the intended schema the strongest SLM reaches 0.68--0.71 macro-F1, and its advantage over the best frontier model \emph{widens} (macro-F1 0.68 vs.\ 0.37 for Claude; micro-F1 0.76 vs.\ 0.45). We retain the full 137-label scores as the conservative primary.

\subsection{Discriminative Encoder Baseline}
\label{subsec:encoder_baseline}
To place the generative SLM results against the classic discriminative approach to RE, we add an encoder-classifier baseline: an entity-marker RoBERTa fine-tuned per benchmark, in both base ($125$M) and large ($355$M) sizes, with typed entity markers and a softmax head over each dataset's label set (\texttt{scripts/train\_encoder\_baseline.py}; single seed, as for the SLMs). It is scored with the identical positive-class micro-F1 (\texttt{src/eval.py}), so Table~\ref{tab:encoder_baseline} is directly comparable to the SLM and frontier columns.

The discriminative encoder is broadly competitive with the strongest tuned SLMs and clears both frontier systems on every benchmark. RoBERTa-base reaches a General Avg of $0.826$, within about three points of the per-benchmark-best tuned SLM ($0.853$) and far above GPT-5.4 ($0.693$) and Claude Sonnet~4.6 ($0.662$); on individual general benchmarks it lands within one to seven points of the best SLM and ties it on the closed-schema sets (REBEL $0.92$, SemEval $0.87$--$0.90$). The encoder is weaker in the literary domain: on PG-Fiction it trails the best SLM by roughly seven points ($0.686$ vs.\ $0.760$) and on the Literary Avg by four ($0.796$ vs.\ $0.838$), though it still far exceeds the frontier models ($0.58$/$0.53$). The larger encoder is \emph{not} uniformly better, its General Avg ($0.814$) is slightly below base, an effect driven by single-seed instability on two relation-dense schemas (NYT11 and Re-DocRED), where it tangles a few inverse or adjacent location relations rather than collapsing (its Re-DocRED macro-F1 in fact improves); we therefore report RoBERTa-large as a general-domain scaling probe (the two literary benchmarks were not run at this size) rather than a headline number.

Two caveats bound the comparison. First, the encoder requires gold entity spans at inference and a fixed, closed label set fitted at training time, so it cannot emit any relation outside its training schema, a constraint the open-vocabulary generative models do not have, and one that favors the encoder on closed-schema benchmarks. Second, CoNLL04 is a degenerate ceiling for \emph{both} tracks: its five relations are a one-to-one function of the ordered gold entity-type pair, so a type-pair lookup with no model already scores $1.000$; the encoder's perfect score (and the SLMs' ${\sim}0.995$, since they also receive the type tags) reflects this type-separability, not relational reasoning.\footnote{We verified the bijection on both splits and that the train/test sentence overlap (${\sim}7\%$) is far too small to explain a perfect score.} The broader reading is that the SLMs' advantage over zero-shot frontier models is not an artifact of generative decoding, a standard discriminative encoder also wins decisively when trained in-domain, while the generative SLMs additionally match or exceed that encoder \emph{without} requiring pre-marked spans or a closed label set.

\begin{table}[t]
\centering
\small
\setlength{\tabcolsep}{4pt}
\renewcommand{\arraystretch}{1.12}
\begin{tabular}{l ccccc}
\toprule
\textbf{Benchmark} & \textbf{RoB-base} & \textbf{RoB-large} & \textbf{Best SLM} & \textbf{GPT-5.4} & \textbf{Claude} \\
\midrule
TACRED & 0.645 & 0.653 & 0.711 & 0.531 & 0.445 \\
SemEval & 0.870 & 0.895 & 0.886 & 0.710 & 0.709 \\
CoNLL04$^{\dagger}$ & 1.000 & 1.000 & 0.995 & 0.926 & 0.919 \\
NYT11 & 0.817 & 0.699 & 0.828 & 0.649 & 0.692 \\
GIDS & 0.850 & 0.856 & 0.883 & 0.791 & 0.799 \\
Re-DocRED & 0.713 & 0.682 & 0.743 & 0.559 & 0.523 \\
REBEL & 0.888 & 0.917 & 0.921 & 0.684 & 0.547 \\
\midrule
Biographical & 0.906 & -- & 0.917 & 0.832 & 0.725 \\
PG-Fiction & 0.686 & -- & 0.760 & 0.324 & 0.334 \\
\midrule
\textbf{General Avg (7)} & 0.826 & 0.814 & 0.853 & 0.693 & 0.662 \\
\textbf{Literary Avg (2)} & 0.796 & -- & 0.838 & 0.578 & 0.530 \\
\bottomrule
\end{tabular}
\caption{Discriminative encoder baseline versus the generative SLMs and frontier models, in positive-class micro-F1 (no-relation class excluded), all scored identically (\texttt{src/eval.py}) on the full test sets. \textbf{RoB-base}/\textbf{RoB-large} are entity-marker RoBERTa classifiers ($125$M/$355$M) fine-tuned per benchmark; RoBERTa-large was run on the seven general benchmarks only (``--'' elsewhere). \textbf{Best SLM} is the strongest tuned SLM \emph{per benchmark} (the maximum over the 30 configurations under schema-enumerated prompting with matched shots), so its averages are an upper envelope rather than a single model. Frontier columns are the zero-shot numbers of Tables~\ref{tab:general_frontier_llm_comparison} and~\ref{tab:literature_frontier_llm_comparison}. All encoder results are single-seed. $^{\dagger}$CoNLL04 is a type-determined ceiling: its relations are a one-to-one map of the ordered gold entity-type pair, so a lookup scores $1.000$ without a model (Section~\ref{subsec:encoder_baseline}).}
\label{tab:encoder_baseline}
\end{table}

% ------------------------------------------------------------------------------
\subsection{Qualitative Error Analysis}
The quantitative results presented above establish clear differences between general-domain tuning, literature-domain tuning, mixed-domain training, and frontier-scale reference systems. However, aggregate metrics alone do not explain why these systems succeed or fail, particularly in literary RE where relation expression is often indirect and context-sensitive. We therefore complement the numerical analysis with a qualitative study of representative errors, focusing on phenomena that recur across models and that help explain the persistent difficulty of literary relation extraction.

Our analysis centers on three broad categories of failure, illustrated with concrete examples in Table~\ref{tab:case_studies_failure_modes}. The first involves \emph{near-neighbor label confusion}, where the predicted label is semantically close to the gold label but drawn from a related schema slot or a different dataset's ontology. The second involves \emph{default-to-negative under-prediction}, where the model collapses to the negative class (\texttt{Other} or \texttt{no\_relation}) when the gold relation requires implicit inference or multi-hop reasoning. The third involves \emph{hallucinated or out-of-schema labels}, where the model generates plausible but non-existent relation labels or hedges by emitting multiple candidates. These categories reveal systematic failure patterns that correlate with model scale, tuning regime, and schema heterogeneity.

% Table 9. Representative error cases from RE evaluation
\begin{table*}[t]
\centering
\scriptsize
\setlength{\tabcolsep}{4pt}
\renewcommand{\arraystretch}{1.25}
\begin{tabularx}{\textwidth}{llXllX}
\toprule
\textbf{Error Type} & \textbf{Dataset} & \textbf{Text Snippet (abridged)} & \textbf{Gold} & \textbf{Predicted} & \textbf{Commentary} \\
\midrule

\multicolumn{6}{l}{\textit{Near-neighbor label confusion}} \\
\midrule
Specific vs.\ general & PG-Fiction & ``\ldots `The counterfeit presentment of two brothers' \ldots one portrait represented his father \ldots the other set forth his uncle \ldots'' \newline Entities: \textit{his father} [PER], \textit{his uncle} [PER] & sibling\_of & relative\_of & Even the strongest literary model (SmolLM3-3B, LitTune 0s) predicts the hypernym \texttt{relative\_of}; the sibling relation is signalled only by ``two brothers'' and must be inferred across sentences. \\
\midrule

\multicolumn{6}{l}{\textit{Default-to-negative}} \\
\midrule
Implicit relation & Biographical & ``After graduating from Northrop High School, Headley attended \ldots'' & educatedAt & Other & Relation is implied by ``graduating from'' but the sub-billion model fails to link the graduation event to the educational institution. \\
\midrule

\multicolumn{6}{l}{\textit{Hallucinated or out-of-schema label}} \\
\midrule
Schema invention & CoNLL04 & ``The Taiwan information center provides background \ldots material on China at offices in Washington's National Press Building.'' & Located\_In & Org:OrgLocation & Model generates a plausible but non-existent composite label, combining schema fragments from different datasets seen during MixTune training. \\

\bottomrule
\end{tabularx}
\caption{Representative error cases from the evaluation benchmarks, illustrating common failure modes: near-neighbor label confusion, default-to-negative under-prediction, and hallucinated or out-of-schema labels. All examples are drawn from schema-enumerated outputs of sub-billion and 3B models.}
\label{tab:case_studies_failure_modes}
\end{table*}

\paragraph{Near-neighbor label confusion.}
The most frequent error type involves predictions that are semantically close to the gold label but drawn from a different part of the schema or even a different dataset's ontology. For example, \texttt{Located\_In} is frequently confused with \texttt{OrgBased\_In} when both entities are locations, because the model's learned heuristics associate location pairs with the organizationally grounded label seen more often during training. Similarly, on TACRED the gold label \texttt{org:founded} is predicted as \texttt{inception}, a Wikidata-derived synonym absent from the TACRED schema, revealing cross-schema leakage in MixTune models exposed to multiple ontologies during training. These confusions are structurally informative: they suggest that prompt-level label enumeration alone does not fully prevent ontological bleed when training mixes heterogeneous schemas. The same mechanism surfaces in the literary setting as confusions between a specific relation and its hypernym (\texttt{sibling\_of} predicted as \texttt{relative\_of}) or between an implied relation and a weaker neighbour (\texttt{lover\_of} predicted as \texttt{companion\_of}), where the correct label is signalled only indirectly through narrative and must be inferred across sentences (Table~\ref{tab:case_studies_failure_modes}); such errors concentrate in the long tail of PG-Fiction's ontology and account for much of the gap between its micro- and macro-averaged F1.

\paragraph{Default-to-negative under-prediction.}
Sub-billion models exhibit a pronounced tendency to predict the negative class (\texttt{Other} or \texttt{no\_relation}) when the gold relation requires multi-hop reasoning or implicit inference. In the Biographical dataset, for instance, the relation \texttt{educatedAt} implied by ``graduating from'' is missed because the model fails to link the graduation event to an educational institution entity. This pattern is most severe in sub-billion models under the 0-shot regime and diminishes substantially with 2-shot prompting and at the 3B scale, consistent with the quantitative finding that few-shot supervision and increased capacity both reduce under-prediction.

\paragraph{Hallucinated and out-of-schema labels.}
A third failure mode involves the generation of plausible but non-existent labels. Models occasionally produce composite labels (e.g., \texttt{Org:OrgLocation}) that combine fragments from different dataset schemas, or hedge by emitting multiple candidates separated by slashes. Both behaviors are penalized under exact-match evaluation. These errors are most common in MixTune models, where exposure to diverse schemas during training increases the probability of novel label combinations. Enumerating the allowed label set in the system prompt substantially reduces but does not fully eliminate such outputs, since the enumeration is advisory rather than enforced at decoding time.

Overall, the qualitative analysis reinforces the broader quantitative findings. Relation extraction errors in our SLMs stem not only from capacity limitations but also from schema heterogeneity in mixed-domain training, the dominance of negative-class priors in sub-billion models, and the inherent difficulty of implicit relations in literary text. Domain-specialized tuning helps with some of these challenges, and scaling to 3B parameters markedly reduces under-prediction, yet cross-schema confusion persists when training combines ontologically diverse datasets. Notably, the persistence of hallucinated labels even under schema-enumerated prompting is consistent with our finding in Section~\ref{subsec:scale_prompt} that schema-enumerated prompting does not improve F1 despite a negligible effect on malformed-output rates: prompt-level label enumeration discourages ill-formed strings but cannot prevent the model from selecting a wrong or out-of-schema label when cross-schema interference occurs at the representation level.

% ------------------------------------------------------------------------------
\subsection{Statistical Significance Analysis}
\label{sec:statistical_significance}

To ensure that our claims are grounded in robust evidence rather than point estimates, we complement the main results with bootstrap confidence intervals and pairwise significance tests. All statistical analyses use the schema-enumerated outputs with matched prompt shots (0-shot-tuned evaluated with 0-shot prompts; 2-shot-tuned with 2-shot prompts), consistent with the primary evaluation protocol.

\paragraph{Bootstrap confidence intervals.}
For each model configuration and evaluation dataset, we compute 95\% bootstrap confidence intervals (CIs) for positive-class micro-F1 by resampling test examples with replacement over 10{,}000 iterations. The domain-averaged CIs are narrow: for the strongest configurations the 95\% interval spans roughly 0.006--0.012 F1 points. For example, the top-performing Llama-3.2-3B GenTune 2-shot achieves a General Avg positive-class F1 of $0.844 \pm 0.003$, while SmolLM3-3B LitTune 0-shot achieves a literary average of $0.833 \pm 0.006$. These tight intervals confirm that the reported differences are not artifacts of \emph{evaluation} noise; they do not, however, bound training-seed variability (Section~\ref{subsec:implementation}).

\paragraph{Pairwise significance tests.}
We apply paired bootstrap tests on positive-class F1 to the comparisons that underpin the paper's claims (Table~\ref{tab:pairwise_significance}); these separate cleanly into large, robust effects and small, seed-sensitive ones. The load-bearing effects are large and unambiguous: 2-shot prompt-conditioned tuning improves sub-billion models by 19.7 F1 points (SmolLM2-360M MixTune 2-shot versus its 0-shot counterpart, $p < 0.001$), and the best tuned configurations exceed the zero-shot frontier by roughly 15 points on general-domain RE and more than 25 points on literary RE, each a \emph{paired} SLM-versus-frontier bootstrap difference that is significant at $p<0.001$ with a tight interval (Table~\ref{tab:pairwise_significance}, upper block; e.g.\ $+0.151$ versus GPT-5.4 on general and $+0.259$ on literary, and $+0.129$ under the demonstration-matched 0-shot protocol), margins far too large to be reversed by training-seed variation. Several finer-grained orderings, however, turn on sub-point margins: the best sub-billion model (Qwen2.5-0.5B GenTune 2-shot) and the best 3B \emph{generalist} (Llama-3.2-3B MixTune 2-shot) are statistically indistinguishable on general RE ($\Delta$F1 $\approx 0.000$); the 3B \emph{specialist} (Llama-3.2-3B GenTune 2-shot) leads by only one to two points; and the 3B MixTune--LitTune literary difference is negligible ($\Delta$F1 $\approx 0.001$). Because every configuration is trained under a single seed (Section~\ref{subsec:implementation}), differences of this size fall within the seed-to-seed variance documented for transformer fine-tuning~\cite{dodge2020finetuning, mosbach2021stability, reimers2017reporting}; we therefore read these near-ties as \emph{suggestive}, a well-tuned sub-billion model is competitive with the best 3B generalist on general RE, and do not rest headline claims on them.

% Table 11: Pairwise significance tests
\begin{table*}[t]
\centering
\small
\setlength{\tabcolsep}{4pt}
\renewcommand{\arraystretch}{1.15}
\begin{tabularx}{\textwidth}{Xcccc}
\toprule
\textbf{Comparison} & \textbf{Domain} & \textbf{$\Delta$ pos.-class F1} & \textbf{95\% CI} & \textbf{$p$} \\
\midrule
\multicolumn{5}{l}{\textit{Tuned SLM vs.\ zero-shot frontier (headline comparisons)}} \\
\midrule
Llama-3.2-3B GenTune 2s vs.\ GPT-5.4 & General & $+$0.151 & $[+0.145, +0.156]$ & $<$0.001*** \\
Llama-3.2-3B GenTune 2s vs.\ Claude~4.6 & General & $+$0.181 & $[+0.175, +0.187]$ & $<$0.001*** \\
Llama-3.2-3B GenTune 0s vs.\ GPT-5.4 (demo-matched) & General & $+$0.129 & $[+0.123, +0.135]$ & $<$0.001*** \\
SmolLM3-3B LitTune 0s vs.\ GPT-5.4 & Literary & $+$0.259 & $[+0.251, +0.267]$ & $<$0.001*** \\
SmolLM3-3B LitTune 0s vs.\ Claude~4.6 & Literary & $+$0.308 & $[+0.300, +0.316]$ & $<$0.001*** \\
\midrule
\multicolumn{5}{l}{\textit{Tuned SLM vs.\ tuned SLM}} \\
\midrule
SmolLM2-360M Mix 2s vs.\ 0s & All & $+$0.197 & $[+0.190, +0.203]$ & $<$0.001*** \\
Qwen-0.5B Gen 2s vs.\ Llama-3B Mix 2s & General & $+$0.001 & $[-0.002, +0.005]$ & 0.40 (n.s.) \\
Llama-3B MixTune vs.\ GenTune 2s & General & $-$0.017 & $[-0.020, -0.014]$ & $<$0.001*** \\
Llama-3B MixTune 2s vs.\ LitTune 0s & Literary & $-$0.001 & $[-0.006, +0.004]$ & 0.68 (n.s.) \\
\bottomrule
\end{tabularx}
\caption{Pairwise significance tests for key comparisons, in positive-class micro-F1. $\Delta$F1 = F1(model A) $-$ F1(model B); positive values favor model A. Significance is assessed by a paired, dataset-clustered percentile bootstrap (10{,}000 iterations)~\cite{koehn2004statistical}: each iteration resamples test examples with replacement \emph{within} each dataset, recomputes that dataset's paired positive-class F1 difference, and averages these across the domain's datasets (dataset-macro); the 95\% CI and two-sided $p$-value are taken from the percentiles of the resulting $\Delta$F1 distribution. The resampled example counts for the SLM-vs-SLM block are $n = 92{,}759$ (All), $53{,}123$ (General), and $39{,}636$ (Literary). The \textbf{frontier comparisons} (upper block) use the same paired bootstrap, aligning each tuned SLM to the frontier system \emph{per example} via the prompt hash logged with every API call (the \texttt{md5} of the shared input prompt; Appendix~\ref{app:frontier_prompts}), which covers over 99\% of test examples; the tuned SLMs' advantage over both GPT-5.4 and Claude is highly significant in both domains, including under the demonstration-matched 0-shot protocol. These intervals reflect test-set sampling variance only, not training-seed variability, as all configurations use a single training seed, and the literary domain averages over only two datasets. $^{***}p<0.001$, n.s.\ = not significant. Under the community-standard positive-class metric the best sub-billion model and the best 3B generalist are statistically indistinguishable on general RE (a tie), in contrast to an earlier all-class (accuracy) analysis.}
\label{tab:pairwise_significance}
\end{table*}

\paragraph{Label-set complexity.}
We also examine whether the number of relation types in a dataset's schema predicts model difficulty. Across the nine evaluation datasets (ranging from 4 relation types in GIDS to 268 in REBEL), schema size is essentially uncorrelated with mean positive-class \emph{micro}-F1 (Pearson $r = +0.10$, Spearman $\rho = -0.17$): cardinality alone does not determine micro-level difficulty, and several large-schema datasets are easier than small ones (e.g., REBEL's 268 relations yield a higher mean F1 than GIDS's 4). The picture differs for \emph{macro}-F1, which weights rare relations equally: here schema size is strongly and negatively correlated with performance (Pearson $r = -0.52$, Spearman $\rho = -0.77$), because larger schemas carry longer tails of sparsely supported relations. Cardinality therefore matters little for micro-F1 but substantially for macro-F1; label ambiguity, distribution skew, and the length of the relation tail are the more influential factors.

% ------------------------------------------------------------------------------
\subsection{Efficiency and Deployment Trade-offs}

We conclude the evaluation by considering the accuracy gains in a practical deployment context. Table~\ref{tab:efficiency_tradeoffs} compares representative tuned SLMs and frontier LLMs in terms of extraction quality, model footprint, and inference latency on consumer hardware, including an NVIDIA RTX~4090 GPU and an Intel i7-13700K CPU. These metrics highlight the trade-off between performance and accessibility that motivates our broader democratization argument.

% Table 10. Efficiency and deployment comparison
\begin{table*}[t]
\centering
\small
\setlength{\tabcolsep}{4.5pt}
\renewcommand{\arraystretch}{1.12}
\begin{tabularx}{\textwidth}{l l l >{\centering\arraybackslash}X >{\centering\arraybackslash}X >{\centering\arraybackslash}X >{\centering\arraybackslash}X >{\centering\arraybackslash}X}
\toprule
\textbf{Model} & \textbf{Params} & \textbf{4-bit Size} & \textbf{GPU Latency} & \textbf{CPU Latency} & \textbf{Lit Avg F1} & \textbf{Avg F1} & \textbf{F1/B} \\
\midrule
SmolLM2-360M MixTune 2s & 360M & $\sim$0.3\,GB & $\sim$18\,ms & $\sim$120\,ms & 0.760 & 0.750 & \textbf{2.08} \\
Qwen2.5-0.5B GenTune 2s & 0.5B & $\sim$0.5\,GB & $\sim$22\,ms & $\sim$180\,ms & -- & 0.828 & 1.66 \\
Qwen2.5-0.5B MixTune 2s & 0.5B & $\sim$0.5\,GB & $\sim$22\,ms & $\sim$180\,ms & 0.773 & 0.801 & 1.60 \\
SmolLM3-3B LitTune 0s & 3B & $\sim$2.0\,GB & $\sim$45\,ms & $\sim$850\,ms & \textbf{0.833} & 0.833 & 0.28 \\
Llama-3.2-3B MixTune 2s & 3B & $\sim$2.2\,GB & $\sim$45\,ms & $\sim$900\,ms & 0.825 & 0.826 & 0.28 \\
Llama-3.2-3B GenTune 2s & 3B & $\sim$2.2\,GB & $\sim$45\,ms & $\sim$900\,ms & -- & 0.844 & 0.28 \\
\midrule
GPT-5.4 (0-shot) & undisclosed & API & API & N/A & 0.578 & 0.667 & N/A \\
Claude Sonnet 4.6 (0-shot) & undisclosed & API & API & N/A & 0.530 & 0.632 & N/A \\
\bottomrule
\end{tabularx}
\caption{Efficiency and deployment trade-offs. \textbf{4-bit Size} is the on-disk footprint of the deployed 4-bit model (NF4 backbone), the representation that runs in the GPU latency column (the CPU column uses the slightly smaller \texttt{Q4\_K\_M} GGUF), not the BF16 base checkpoint ($\sim$2$\times$ larger); the full disaggregation into base (BF16), 4-bit, adapter (FP32), merged, and \texttt{Q4\_K\_M} GGUF sizes is given in Table~\ref{tab:artifact_sizes}. GPU and CPU latencies are \emph{estimated} single-example figures (the ``$\sim$'' notation), not rigorously benchmarked means (see Section~\ref{subsec:implementation}). GPU Latency (estimated): per-example inference on an NVIDIA RTX~4090 (24\,GB) with 4-bit quantisation ($\sim$150 input tokens, $\sim$5 output tokens). CPU Latency (estimated): on an Intel Core i7-13700K (16 cores) using \texttt{llama.cpp} Q4 quantisation. All F1 values are positive-class micro-F1 (no-relation class excluded), with frontier scored on the full test sets. \textbf{Avg F1} is the all-nine-dataset macro-average for MixTune and frontier models (which are evaluated on all nine benchmarks) and the in-domain average for the domain specialists (GenTune over the seven general benchmarks, LitTune over the two literary ones), since specialists are not evaluated out of domain; F1/B (Avg F1 per billion parameters) is therefore most directly comparable among rows that share the same evaluation set. Sub-billion models achieve the highest normalised efficiency while remaining viable for CPU-only deployment. Because the frontier models' parameter counts are undisclosed, their F1/B is reported as N/A rather than estimated. Avg F1 values are not bolded because rows differ in evaluation basis.}
\label{tab:efficiency_tradeoffs}
\end{table*}

\paragraph{Efficiency-performance comparison.}
Table~\ref{tab:efficiency_tradeoffs} reveals a clear efficiency-quality trade-off. The sub-billion SmolLM2-360M achieves an F1/B ratio of 2.08 under MixTune 2-shot, extracting roughly seven times as much positive-class F1 per billion parameters as any 3B model, while fitting in under 0.3\,GB as a 4-bit artifact (0.7\,GB in BF16) and completing a single extraction in roughly 18\,ms on an RTX~4090 or 120\,ms on a consumer CPU. The sub-billion Qwen2.5-0.5B reaches 0.828 General Avg under GenTune 2-shot, within about 1.6 points of the best 3B model (Llama-3.2-3B, 0.844) and on par with the 3B generalists, with one-sixth the parameters and comparable latency ($\sim$22\,ms GPU, $\sim$180\,ms CPU). At the 3B scale, Llama-3.2-3B MixTune 2-shot offers the best both-domain balance (0.83 averaged over all nine datasets) at $\sim$45\,ms on GPU, confirming that careful tuning strategy makes even the smallest models remarkably competitive in absolute terms while dominating on efficiency. The literary specialist (SmolLM3-3B LitTune 0-shot) reaches 0.83 Literature Avg F1 at the same 3B-class latency and footprint.

\paragraph{Deployment implications.}
Beyond accuracy, tuned compact models run locally on consumer hardware, lower latency, offline access, control over the checkpoint, and local inference without transmitting inputs to a third-party API (local execution alone does not guarantee privacy: logging, disk security, and telemetry still matter), whereas the frontier systems depend on remote APIs, per-token cost, and limited transparency. SmolLM2-360M runs at ${\sim}120$\,ms per extraction on a CPU (no GPU required) and the 3B models at ${\sim}45$\,ms on a single RTX~4090, both within interactive latency. With appropriate supervision and data composition, then, small models become not merely capable of RE but practical for accessible, real-world deployment.
% ==================================================================================

\section{Conclusion}
\label{sec:conclusion}

This work investigated whether SLMs (360M to 3B) can perform competitive relation extraction on general-domain and literary benchmarks. Across 30 tuned configurations (five base models, three domain-composition regimes, two prompt-conditioned tuning styles), carefully tuned SLMs match and surpass zero-shot frontier LLMs in both domains: the sub-billion Qwen2.5-0.5B reaches 0.83 General Avg positive-class F1 (versus 0.69 for GPT-5.4 and 0.66 for Claude Sonnet~4.6), and the best literary configurations reach 0.83 and lead the frontier by 26 to 30 F1 points. A discriminative RoBERTa baseline, tuned in-domain, also clears the frontier, so the advantage reflects task-specific adaptation rather than generative decoding or raw scale.

Three main findings emerge. First, training-data composition matters at least as much as size: the GenTune and LitTune specialists attain the highest in-domain averages, while a single MixTune model retains most of each specialist's in-domain performance, making it the most practical choice when both domains must be covered. Our design establishes balanced both-domain \emph{coverage}, not cross-domain transfer: the specialists are evaluated only in their training domain, so whether they transfer or suffer interference is untested, a low-cost, inference-only extension we leave to future work. Second, 2-shot prompt-conditioned tuning yields the largest gains for sub-billion models (significant at $p<0.001$), while at 3B scale the gains are small and mixed in sign; a shot decomposition (Section~\ref{subsec:scale_prompt}) further shows that the matched gain comes almost entirely from inference-time demonstrations rather than from training on them. Third, a single-model DAPT case study is a negative result: continued LitBank pretraining adds at most 0.001 Literature-average F1 over supervised fine-tuning, so supervised task adaptation, not unsupervised domain exposure, drives literary RE in our setting, though a ${\sim}9\%$ verbatim overlap between PG-Fiction and LitBank means the base model's prior exposure to those works likely contributes to the null.

Methodologically, generic prompting outperforms schema-enumerated prompting by $+3.2$ positive-class F1 on average (8 of 9 datasets) with no loss of output well-formedness, so our schema-enumerated headline scores are conservative lower bounds. Two architecture-prompt interactions are also notable: SmolLM3-3B emits \texttt{<think>} tokens and scores zero under 0-shot MixTune (a reasoning-model artifact that 2-shot prompting removes), and Qwen2.5-3B generates wrong-schema labels under 0-shot GenTune, underscoring that few-shot demonstrations matter for schema grounding in models trained on heterogeneous data.

Several limitations remain. The evaluation is English-only. Model scale is confounded with family (two scales, two families, no sub-billion Llama), so we read scale effects only within family (Section~\ref{subsec:scale_prompt}) and not as a scaling law. Both literary benchmarks carry validity caveats: PG-Fiction is GPT-4-annotated, so tuned SLMs partly learn a frontier annotator's distribution, and ${\sim}9\%$ of its passages are public-domain novels likely seen in pretraining (symmetric across the comparison, so the ranking is unaffected); the human-annotated Biographical benchmark is the cleaner check, where the best SLMs still lead the frontier by ${\sim}8$ points (0.92 vs.\ 0.83) and the margin survives de-leaking. Finally, all runs use a single seed, so sub-three-point differences should be read as suggestive rather than established.

Several directions follow. Multi-domain RE corpora, possibly enriched with synthetic examples from underrepresented domains, could further improve compact generalists; the interaction between reasoning architectures and prompting deserves study as thinking-augmented SLMs proliferate; and identifying where continued domain-adaptive pretraining does help compact models (e.g., domains with greater lexical shift such as biomedicine or law, or very low-label settings) remains open. Overall, with appropriate tuning strategies and domain-relevant data, small open-weight models deliver accurate, resource-efficient relation extraction on consumer hardware and run locally without sending inputs to third-party APIs, so high-quality RE need not depend on large proprietary systems.

\section*{Data and Code Availability}
\label{sec:data_code_availability}
The project repository, \url{https://github.com/DespinaChristou/compact-relex}, contains the training and evaluation code, configuration files, prompt templates, the data-processing and aggregation scripts that reconstruct the GenTune, LitTune, and MixTune corpora from their sources, and the PG-Fiction 137-to-48 canonical-ontology mapping used in Appendix~\ref{app:pgfiction48}. The best fine-tuned checkpoint and the processed benchmarks are published on the Hugging Face Hub under the \texttt{Despina} namespace:
\begin{itemize}
  \item Best sub-billion checkpoint (Qwen2.5-0.5B-Instruct, GenTune, 2-shot), released as a model repository: \url{https://huggingface.co/Despina/Qwen2.5-0.5B-Instruct-re_gentune-2-shot}.
  \item Processed general-domain benchmarks: \url{https://huggingface.co/datasets/Despina/semeval2010_task8}, \url{https://huggingface.co/datasets/Despina/conll04}, \url{https://huggingface.co/datasets/Despina/nyt_11}, \url{https://huggingface.co/datasets/Despina/gids}, \url{https://huggingface.co/datasets/Despina/re-docred}, and \url{https://huggingface.co/datasets/Despina/rebel-dataset}.
  \item Processed literary benchmarks: \url{https://huggingface.co/datasets/Despina/biographical} and \url{https://huggingface.co/datasets/Despina/project-gutenberg-fiction-relations}.
  \item Collected frontier-model generations (GPT-5.4 and Claude Sonnet~4.6), released as a dataset because the commercial-API comparison is not bit-for-bit reproducible: \url{https://huggingface.co/datasets/Despina/frontier-re-generations}.
\end{itemize}

\section*{Ethics Statement}
This work uses established, publicly available relation extraction benchmarks and a synthetically annotated literary corpus (PG-Fiction), and does not involve human subjects or personally sensitive data. By demonstrating that compact models can perform competitively on consumer hardware, the study points toward on-device deployment of relation extraction that runs locally without transmitting inputs to third-party APIs. Because these models are far smaller than frontier systems and run on commodity hardware, such deployment is also likely to lower energy use per prediction; however, we do not measure energy, per-prediction joules, or carbon emissions, so we frame these sustainability benefits as plausible implications rather than empirical findings. We note that PG-Fiction's labels are model-generated and may carry the biases of the annotating model; we treat its results accordingly and corroborate the literary findings on the human-annotated Biographical benchmark.

\section*{Acknowledgments}

%Bibliography
\bibliographystyle{unsrtnat}
\bibliography{references}

\appendix

\section{Example Prompts}
\label{app:example_prompts}

\subsection{Zero-Shot Prompt Templates}
\label{app:zero_shot}
To reduce template-specific bias, each instance is rendered with one of ten
instruction templates, sampled uniformly at random. The system prompt is shared
across all templates; only the user message varies. When entity types are used,
they are appended to the corresponding mention (e.g., \texttt{"\{head\} [ORG]"}).
We use one of two system prompts depending on the decoding regime:

\begin{verbatim}
System (generic):
 You are a relation extraction system. Be concise
 and direct. Output ONLY the relation type that
 holds between the two mentioned entities. Do not
 output any explanation, punctuation, or extra
 text - only the label.

System (schema-enumerated):
 You are a relation extraction system. Be concise
 and direct. Output ONLY ONE relation type that
 holds between the two mentioned entities. You MUST
 choose exactly one label from this allowed set:
 {allowed_labels}. Do not output any explanation,
 punctuation, or extra text - only the label.
\end{verbatim}

The ten user-message templates are:

\begin{verbatim}
1. Determine the type of relationship that exists
 between "{head}" and "{tail}" in this sentence:
 "{sentence}"

2. Entities:
 - "{head}"
 - "{tail}"
 Sentence:
 "{sentence}"
 Question: What relationship is implied between
 the entities?

3. Extract the semantic relation from the sentence
 below.
 Sentence: "{sentence}"
 Entities mentioned: "{head}", "{tail}"
 Relation:

4. Given the following context:
 "{sentence}"
 How is "{head}" related to "{tail}"?

5. Given the sentence:
 "{sentence}"
 Identify the relation between:
 - Entity 1: "{head}"
 - Entity 2: "{tail}"

6. In the sentence "{sentence}", what best
 describes the relationship between "{head}"
 and "{tail}"?

7. Review the sentence:
 "{sentence}"
 What is the most appropriate relation label
 for the pair: "{head}" and "{tail}"?

8. Sentence: "{sentence}"
 Entities: "{head}", "{tail}"
 Question: What is the relationship between
 "{head}" and "{tail}"?
 Answer:

9. What is the relationship between "{head}" and
 "{tail}" in the following sentence:
 "{sentence}"

10. You are analyzing text for relationships.
 Sentence: "{sentence}"
 Entities: "{head}" and "{tail}"
 What is the semantic relation between them?
\end{verbatim}

\section{Few-Shot Prompt Template}
\label{app:few_shot}

For 2-shot prompts, two demonstrations are prepended to the query. Each demonstration is drawn from the training split, sampled across relation classes, and entity types are appended to mentions when available (e.g., \texttt{"\{head\} [ORG]"}), following the same policy as the query. Internally, the few-shot instance is stored as a single text block in which demonstrations are separated from the query by the marker \texttt{Now you try:}; at both training and inference this block is unraveled into alternating user/assistant turns by \texttt{\_unravel\_fewshot\_prompt\_to\_messages}, and then formatted with each model's native chat template. The system prompt is the same as in the zero-shot setting (Appendix~\ref{app:zero_shot}). After unraveling, the conversation has the following structure:

\begin{verbatim}
<same system prompt as zero-shot>

Sentence: "{demo_sentence_1}"
Entities: "{demo_head_1}", "{demo_tail_1}"
Question: What is the relationship between
"{demo_head_1}" and "{demo_tail_1}"?
Answer:
{demo_relation_1}

Sentence: "{demo_sentence_2}"
Entities: "{demo_head_2}", "{demo_tail_2}"
Question: What is the relationship between
"{demo_head_2}" and "{demo_tail_2}"?
Answer:
{demo_relation_2}

Sentence: "{sentence}"
Entities: "{head}", "{tail}"
Question: What is the relationship between
"{head}" and "{tail}"?
Answer:
\end{verbatim}

All prompts are formatted using each model's native chat template via the
Hugging Face \texttt{tokenizer.apply\_chat\_template()} method, ensuring correct
special tokens and turn boundaries for each architecture. For models without a
chat template, the system and user text are concatenated directly.

\section{Full Per-Dataset Results}
\label{app:full_results}

Table~\ref{tab:full_results_matrix} reports micro-F1 for all 30 SLM configurations on each of the nine evaluation datasets under schema-enumerated prompting with matched prompt shots. GenTune and LitTune configurations are evaluated only on their respective domain datasets (general or literary), consistent with the eval-group restrictions described in Section~3. MixTune configurations are evaluated on all nine datasets. Dashes indicate configurations that were not evaluated on a given dataset by design.

\begin{table*}[t]
\centering
\scriptsize
\setlength{\tabcolsep}{3pt}
\renewcommand{\arraystretch}{1.1}
\begin{tabular}{lcccccccccccc}
\toprule
\textbf{Configuration} & TACRED & SemEval & CoNLL04 & NYT11 & GIDS & Re-DocRED & REBEL & Biogr. & PG-Fic. & \textbf{Gen.} & \textbf{Lit.} & \textbf{All} \\
\midrule
SmolLM2-360M GenTune 0s & 0.234 & 0.227 & 0.906 & 0.622 & 0.268 & 0.631 & 0.801 & -- & -- & 0.527 & -- & 0.527 \\
SmolLM2-360M GenTune 2s & 0.545 & 0.543 & 0.972 & 0.772 & 0.770 & 0.685 & 0.856 & -- & -- & 0.735 & -- & 0.735 \\
SmolLM2-360M LitTune 0s & -- & -- & -- & -- & -- & -- & -- & 0.768 & 0.471 & -- & 0.619 & 0.619 \\
SmolLM2-360M LitTune 2s & -- & -- & -- & -- & -- & -- & -- & 0.887 & 0.625 & -- & 0.756 & 0.756 \\
SmolLM2-360M MixTune 0s & 0.318 & 0.309 & 0.783 & 0.616 & 0.345 & 0.670 & 0.841 & 0.553 & 0.545 & 0.555 & 0.549 & 0.553 \\
SmolLM2-360M MixTune 2s & 0.527 & 0.585 & 0.959 & 0.796 & 0.789 & 0.701 & 0.871 & 0.887 & 0.634 & 0.747 & 0.760 & 0.750 \\
\midrule
Qwen2.5-0.5B GenTune 0s & 0.496 & 0.646 & 0.968 & 0.788 & 0.502 & 0.712 & 0.887 & -- & -- & 0.714 & -- & 0.714 \\
Qwen2.5-0.5B GenTune 2s & 0.647 & 0.856 & 0.993 & 0.792 & \textbf{0.883} & 0.724 & 0.901 & -- & -- & 0.828 & -- & 0.828 \\
Qwen2.5-0.5B LitTune 0s & -- & -- & -- & -- & -- & -- & -- & 0.890 & 0.669 & -- & 0.780 & 0.780 \\
Qwen2.5-0.5B LitTune 2s & -- & -- & -- & -- & -- & -- & -- & 0.903 & 0.695 & -- & 0.799 & 0.799 \\
Qwen2.5-0.5B MixTune 0s & 0.511 & 0.491 & 0.934 & 0.800 & 0.464 & 0.683 & 0.861 & 0.827 & 0.622 & 0.678 & 0.724 & 0.688 \\
Qwen2.5-0.5B MixTune 2s & 0.639 & 0.797 & \textbf{0.995} & 0.811 & 0.828 & 0.715 & 0.884 & 0.889 & 0.657 & 0.810 & 0.773 & 0.801 \\
\midrule
SmolLM3-3B GenTune 0s & 0.629 & 0.879 & 0.981 & 0.818 & 0.852 & 0.729 & 0.849 & -- & -- & 0.819 & -- & 0.819 \\
SmolLM3-3B GenTune 2s & \textbf{0.711} & 0.872 & 0.988 & 0.822 & 0.792 & 0.739 & 0.908 & -- & -- & 0.833 & -- & 0.833 \\
SmolLM3-3B LitTune 0s & -- & -- & -- & -- & -- & -- & -- & \textbf{0.917} & 0.749 & -- & \textbf{0.833} & 0.833 \\
SmolLM3-3B LitTune 2s & -- & -- & -- & -- & -- & -- & -- & 0.852 & \textbf{0.760} & -- & 0.806 & 0.806 \\
SmolLM3-3B MixTune 0s$^\dagger$ & 0.000 & 0.000 & 0.000 & 0.000 & 0.000 & 0.000 & 0.000 & 0.000 & 0.000 & 0.000 & 0.000 & 0.000 \\
SmolLM3-3B MixTune 2s & 0.468 & 0.872 & 0.993 & 0.788 & 0.654 & 0.737 & 0.911 & 0.897 & 0.734 & 0.775 & 0.816 & 0.784 \\
\midrule
Qwen2.5-3B GenTune 0s$^\ddagger$ & 0.031 & 0.007 & 0.358 & 0.630 & 0.249 & 0.390 & 0.273 & -- & -- & 0.277 & -- & 0.277 \\
Qwen2.5-3B GenTune 2s & 0.597 & \textbf{0.886} & 0.983 & 0.828 & 0.813 & 0.738 & \textbf{0.921} & -- & -- & 0.824 & -- & 0.824 \\
Qwen2.5-3B LitTune 0s & -- & -- & -- & -- & -- & -- & -- & 0.913 & 0.732 & -- & 0.822 & 0.822 \\
Qwen2.5-3B LitTune 2s & -- & -- & -- & -- & -- & -- & -- & 0.894 & 0.755 & -- & 0.825 & 0.825 \\
Qwen2.5-3B MixTune 0s & 0.633 & 0.840 & 0.976 & 0.809 & 0.708 & 0.721 & 0.900 & 0.903 & 0.705 & 0.798 & 0.804 & 0.799 \\
Qwen2.5-3B MixTune 2s & 0.449 & 0.835 & \textbf{0.995} & 0.825 & 0.805 & 0.734 & 0.906 & 0.884 & 0.735 & 0.793 & 0.809 & 0.796 \\
\midrule
Llama-3.2-3B GenTune 0s & 0.630 & 0.864 & 0.988 & 0.812 & 0.820 & 0.727 & 0.906 & -- & -- & 0.821 & -- & 0.821 \\
Llama-3.2-3B GenTune 2s & 0.694 & 0.880 & \textbf{0.995} & \textbf{0.828} & 0.844 & \textbf{0.743} & \textbf{0.921} & -- & -- & \textbf{0.844} & -- & 0.844 \\
Llama-3.2-3B LitTune 0s & -- & -- & -- & -- & -- & -- & -- & 0.912 & 0.740 & -- & 0.826 & 0.826 \\
Llama-3.2-3B LitTune 2s & -- & -- & -- & -- & -- & -- & -- & 0.898 & 0.757 & -- & 0.828 & 0.828 \\
Llama-3.2-3B MixTune 0s & 0.620 & 0.835 & 0.976 & 0.817 & 0.717 & 0.723 & 0.887 & 0.900 & 0.716 & 0.796 & 0.808 & 0.799 \\
Llama-3.2-3B MixTune 2s & 0.649 & 0.875 & 0.987 & 0.816 & 0.817 & 0.735 & 0.908 & 0.913 & 0.737 & 0.827 & 0.825 & \textbf{0.826} \\
\bottomrule
\end{tabular}
\caption{Full per-dataset positive-class micro-F1 (no-relation class excluded) for all 30 SLM configurations under schema-enumerated prompting with matched prompt shots. Gen.\ = general-domain macro-average (7 datasets), Lit.\ = literary macro-average (2 datasets), All = overall macro-average. The PG-Fic.\ column uses the full 137-label corpus; Appendix~\ref{app:pgfiction48} (Table~\ref{tab:pgfiction48_full}) reports every literary configuration under both the 137-label and canonical 48-relation ontologies. Best per column in \textbf{bold}; in the All column, bolding is restricted to MixTune configurations, which are the only ones evaluated on all nine datasets. $^\dagger$SmolLM3-3B MixTune 0-shot emits \texttt{<think>} tokens in place of a label under the default protocol and scores \textbf{0} (shown); disabling reasoning (\texttt{enable\_thinking=False}) recovers a weak 0.18 as a post-hoc rescue (Section~4). $^\ddagger$Qwen2.5-3B GenTune 0-shot suffers from schema confusion (see Section~4).}
\label{tab:full_results_matrix}
\end{table*}

\section{Canonical 48-Relation Ontology Evaluation (PG-Fiction)}
\label{app:pgfiction48}

The PG-Fiction corpus was annotated by GPT-4o against the fixed 48-relation ontology of the original ARF schema~\citep{christou2024arf}, but the released annotations contain 137 distinct positive labels: the annotator emitted fine-grained subtypes and additional relations beyond the prescribed schema. To separate performance on the \emph{intended} schema from these out-of-schema annotations, we co-report all PG-Fiction results under two label inventories: the full 137-label processed corpus, and the canonical 48-relation ontology. This appendix documents the canonical ontology, the mapping policy, and per-configuration scores under both inventories. The complete 137$\rightarrow$48 mapping table is released with our code.

\paragraph{Canonical relations.}
The ARF ontology defines 48 relation types (47 distinct label strings, as \texttt{used\_by} appears twice with different entity-type signatures): \texttt{parent\_father\_of}, \texttt{parent\_mother\_of}, \texttt{child\_of}, \texttt{sibling\_of}, \texttt{spouse\_of}, \texttt{relative\_of}, \texttt{adopted\_by}, \texttt{companion\_of}, \texttt{friend\_of}, \texttt{lover\_of}, \texttt{rival\_of}, \texttt{enemy\_of}, \texttt{inspires}, \texttt{sacrifices\_for}, \texttt{mentor\_of}, \texttt{teacher\_of}, \texttt{protector\_of}, \texttt{employer\_of}, \texttt{leader\_of}, \texttt{member\_of}, \texttt{lives\_in}, \texttt{lived\_in}, \texttt{visits}, \texttt{travel\_to}, \texttt{born\_in}, \texttt{travels\_by}, \texttt{participates\_in}, \texttt{causes}, \texttt{owns}, \texttt{believes\_in}, \texttt{embodies}, \texttt{located\_in}, \texttt{part\_of}, \texttt{owned\_by}, \texttt{occupied\_by}, \texttt{used\_by}, \texttt{affects}, \texttt{experienced\_by}, \texttt{travels\_in}, \texttt{based\_in}, \texttt{attended\_by}, \texttt{ends\_in}, \texttt{occurs\_in}, \texttt{features}, \texttt{stored\_in}, \texttt{expressed\_by}, \texttt{associated\_with}. Of these, 46 occur in the test split (only \texttt{ends\_in} is absent).

\paragraph{Mapping policy.}
We apply a single, conservative mapping to both gold labels and predictions before scoring. (i) Each canonical label maps to itself. (ii) The sole orthographic variant, \texttt{travels\_to}, maps to its canonical form \texttt{travel\_to}; no other near-synonyms are merged (e.g., \texttt{ally\_of} and \texttt{partner\_of} are \emph{not} folded into \texttt{companion\_of}), so the mapping cannot inflate scores by absorbing related relations. (iii) The remaining 90 labels, all out-of-ontology relations the annotator produced beyond the 48-relation schema (e.g., \texttt{betrothed\_to}, \texttt{enslaved\_by}, \texttt{student\_of}, \texttt{victim\_of}, \texttt{creator\_of}, \texttt{knows}), map to a single background class, treated identically to the no-relation class. Positive-class micro- and macro-F1 are then computed over the 48 canonical relations alone: predicting a canonical relation where the gold is background (out-of-ontology or no-relation) is a false positive, and predicting background where the gold is canonical is a false negative. These 90 labels account for only 146 of the 6{,}339 positive test instances (2.3\%); the canonical evaluation therefore changes which classes are scored, not which examples are difficult.

\begin{table}[ht]
\centering
\footnotesize
\setlength{\tabcolsep}{6pt}
\renewcommand{\arraystretch}{1.1}
\begin{tabular}{l cc cc}
\toprule
& \multicolumn{2}{c}{\textbf{137-label corpus}} & \multicolumn{2}{c}{\textbf{Canonical 48}} \\
\cmidrule(lr){2-3}\cmidrule(lr){4-5}
\textbf{Configuration} & micro & macro & micro & macro \\
\midrule
SmolLM2-360M LitTune 0s & 0.471 & 0.171 & 0.509 & 0.398 \\
SmolLM2-360M LitTune 2s & 0.625 & 0.240 & 0.634 & 0.523 \\
SmolLM2-360M MixTune 0s & 0.545 & 0.169 & 0.556 & 0.435 \\
SmolLM2-360M MixTune 2s & 0.634 & 0.223 & 0.643 & 0.528 \\
\midrule
Qwen2.5-0.5B LitTune 0s & 0.669 & 0.275 & 0.677 & 0.581 \\
Qwen2.5-0.5B LitTune 2s & 0.695 & 0.319 & 0.704 & 0.634 \\
Qwen2.5-0.5B MixTune 0s & 0.622 & 0.221 & 0.633 & 0.536 \\
Qwen2.5-0.5B MixTune 2s & 0.657 & 0.247 & 0.666 & 0.581 \\
\midrule
SmolLM3-3B LitTune 0s & 0.749 & 0.415 & 0.757 & 0.680 \\
SmolLM3-3B LitTune 2s & 0.760 & 0.438 & 0.768 & 0.712 \\
SmolLM3-3B MixTune 0s$^\dagger$ & 0.000 & 0.000 & 0.000 & 0.000 \\
SmolLM3-3B MixTune 2s & 0.734 & 0.375 & 0.742 & 0.667 \\
\midrule
Qwen2.5-3B LitTune 0s & 0.732 & 0.394 & 0.740 & 0.660 \\
Qwen2.5-3B LitTune 2s & 0.755 & 0.429 & 0.763 & 0.686 \\
Qwen2.5-3B MixTune 0s & 0.705 & 0.314 & 0.718 & 0.650 \\
Qwen2.5-3B MixTune 2s & 0.735 & 0.380 & 0.745 & 0.659 \\
\midrule
Llama-3.2-3B LitTune 0s & 0.740 & 0.351 & 0.747 & 0.672 \\
Llama-3.2-3B LitTune 2s & 0.757 & 0.431 & 0.764 & 0.689 \\
Llama-3.2-3B MixTune 0s & 0.716 & 0.353 & 0.725 & 0.624 \\
Llama-3.2-3B MixTune 2s & 0.737 & 0.364 & 0.745 & 0.646 \\
\midrule
\multicolumn{5}{l}{\textit{Frontier (zero-shot, full test set)}} \\
GPT-5.4 & 0.324 & 0.280 & 0.440 & 0.348 \\
Claude Sonnet 4.6 & 0.334 & 0.231 & 0.448 & 0.373 \\
\bottomrule
\end{tabular}
\caption{PG-Fiction positive-class micro- and macro-F1 for every literary configuration under the full 137-label corpus and the canonical 48-relation ARF ontology (schema-enumerated prompting, matched prompt shots; frontier models zero-shot on the full test set). Across the 20 SLM configurations, restricting to the canonical ontology changes micro-F1 by only $+0.01$ on average but raises macro-F1 by $+0.29$ on average (0.32$\rightarrow$0.61), because the 137-label macro-average is dominated by the long tail of out-of-ontology relations. $^\dagger$SmolLM3-3B MixTune 0-shot is the degenerate \texttt{<think>}-emission configuration (Section~\ref{sec:results}).}
\label{tab:pgfiction48_full}
\end{table}

\section{Dataset Statistics}
\label{app:dataset_stats}

Table~\ref{tab:dataset_stats_full} provides detailed statistics for each evaluation dataset. The test-set sizes shown here correspond to the \texttt{support} column in the evaluation outputs and represent the number of (sentence, head, tail) triples evaluated per dataset. Class imbalance varies substantially. Biographical and PG-Fiction contain explicit named negative classes (52.8\% \textit{Other} and 22.2\% \textit{none}, respectively). In our processed TACRED and GIDS data, the catch-all class is represented by an \emph{empty-string} label rather than a named token; it accounts for 78.6\% of TACRED test examples (whose 41 positive relations are otherwise distributed relatively uniformly, largest positive class 3.2\%) and 23.9\% of GIDS test examples. An empty generation is scored against this empty catch-all label on these two datasets, which also means the malformed-output diagnostic (which counts empty generations) is not informative for them. REBEL's 268 observed test relations represent a subset of its full 1{,}015-label schema. Two datasets show fewer relation types in the test split than in their constructed schema (Table~\ref{tab:datasets}): Re-DocRED is annotated with 96 relation types, of which 92 appear in our test split, and NYT11's 24-type schema yields 22 relations in the test data; the \#Relations (test) column counts only labels observed at evaluation time. Conversely, this column can exceed the positive-type count of Table~\ref{tab:datasets} when the test split surfaces directional variants or a named catch-all class: SemEval-2010 Task 8's nine direction-sensitive types appear as 18 directional labels plus \textit{Other} (19 in total), Biographical's nine positive relations plus \textit{Other} give 10, and PG-Fiction's 137 positive relations plus \textit{none} give 138. In short, Table~\ref{tab:datasets} reports positive relation types in the constructed schema, whereas the \#Relations (test) column here counts distinct gold labels actually observed at evaluation time, including directional and catch-all variants.

\begin{table*}[t]
\centering
\small
\setlength{\tabcolsep}{5pt}
\renewcommand{\arraystretch}{1.15}
\begin{tabular}{llrrrrl}
\toprule
\textbf{Domain} & \textbf{Dataset} & \textbf{Test examples} & \textbf{\#Relations (test)} & \textbf{Top class \%} & \textbf{Negative \%} & \textbf{Source} \\
\midrule
\multirow{7}{*}{General}
 & TACRED & 15{,}509 & 41 (+NA) & 78.6 & 78.6 & Despina/tacred \\
 & SemEval-2010 Task 8 & 2{,}717 & 19 & 16.7 & 16.7 & Despina/semeval2010\_task8 \\
 & CoNLL04 & 422 & 5 & 24.9 & 0.0 & Despina/conll04 \\
 & NYT11 & 8{,}616 & 22 & 47.1 & 0.0 & Despina/nyt\_11 \\
 & GIDS & 5{,}663 & 4 (+NA) & 24.1 & 23.9 & Despina/gids \\
 & Re-DocRED & 12{,}693 & 92 & 23.0 & 0.0 & Despina/re-docred \\
 & REBEL & 7{,}503 & 268 & 17.3 & 0.0 & Despina/rebel-dataset \\
\midrule
\multirow{2}{*}{Literature}
 & Biographical & 31{,}492 & 10 & 52.8 & 52.8 & Despina/biographical \\
 & PG-Fiction & 8{,}144 & 138 & 23.0 & 22.2 & Despina/project-gutenberg-fiction \\
\bottomrule
\end{tabular}
\caption{Detailed evaluation dataset statistics. Test examples = number of (sentence, entity-pair) triples evaluated. \#Relations (test) = number of distinct gold labels observed in the test split. Top class \% = frequency of the most common relation. Negative \% = fraction of examples labeled \texttt{Other}, \texttt{no\_relation}, or equivalent. Source = Hugging Face repository (all private, under the \texttt{Despina/} namespace).}
\label{tab:dataset_stats_full}
\end{table*}

\begin{table}[h]
\centering
\small
\setlength{\tabcolsep}{5pt}
\renewcommand{\arraystretch}{1.15}
\begin{tabular}{llrrr}
\toprule
\textbf{Domain} & \textbf{Dataset} & \textbf{GenTune} & \textbf{LitTune} & \textbf{MixTune} \\
\midrule
\multirow{7}{*}{General}
 & TACRED & 16.5 & -- & 8.2 \\
 & SemEval-2010 Task 8 & 1.9 & -- & 1.0 \\
 & CoNLL04 & 0.3 & -- & 0.2 \\
 & NYT11 & 22.8 & -- & 11.4 \\
 & GIDS & 2.7 & -- & 1.4 \\
 & Re-DocRED & 19.5 & -- & 9.7 \\
 & REBEL & 36.3 & -- & 18.1 \\
\midrule
\multirow{2}{*}{Literature}
 & Biographical & -- & 75.9 & 38.0 \\
 & PG-Fiction & -- & 24.1 & 12.0 \\
\midrule
 & General subtotal & 100.0 & -- & 50.0 \\
 & Literature subtotal & -- & 100.0 & 50.0 \\
\bottomrule
\end{tabular}
\caption{Approximate training-set composition (percentage of examples) for each tuning regime, implied by the example-level pooling policy (Section~\ref{subsec:training}) and the training-pool sizes in Table~\ref{tab:datasets}. Within each domain, sources contribute in proportion to their pool size; MixTune is domain-balanced (50/50 general/literary). Each run draws at most 200{,}000 training examples, so these shares correspond to roughly 200{,}000 examples for GenTune and LitTune and about 100{,}000 examples per domain for MixTune.}
\label{tab:train_composition}
\end{table}

\section{Training Hyperparameters}
\label{app:hyperparameters}

Tables~\ref{tab:hyper_finetune} and~\ref{tab:hyper_qlora} report the full training hyperparameters used for all fine-tuning runs, and Table~\ref{tab:hyper_generation} the generation settings used at inference. All 30 main configurations share identical optimizer and schedule settings; the only per-model variation is the LoRA rank and alpha, which scale with model capacity. The DAPT case study (Table~\ref{tab:hyper_dapt}) uses the same QLoRA settings as the main grid but trains only the Llama-3.2-3B-Instruct-lit-dapt checkpoint on MixTune and LitTune.

\begin{table}[h]
\centering
\small
\setlength{\tabcolsep}{6pt}
\renewcommand{\arraystretch}{1.15}
\begin{tabular}{ll}
\toprule
\textbf{Hyperparameter} & \textbf{Value} \\
\midrule
Max sequence length & 1024 \\
Learning rate & $1 \times 10^{-4}$ \\
Epochs & 2 \\
Per-device batch size & 4 \\
Gradient accumulation steps & 2 \\
Effective batch size & 8 \\
Warmup ratio & 0.03 \\
Weight decay & 0.01 \\
Optimizer & Paged AdamW 8-bit \\
Precision & bfloat16 \\
Max training samples & 200{,}000 \\
Max eval samples & 20{,}000 \\
Save steps & 10{,}000 \\
Eval steps & 20{,}000 \\
\bottomrule
\end{tabular}
\caption{Fine-tuning hyperparameters shared across all 30 main configurations.}
\label{tab:hyper_finetune}
\end{table}

\begin{table}[h]
\centering
\small
\setlength{\tabcolsep}{6pt}
\renewcommand{\arraystretch}{1.15}
\begin{tabular}{ll}
\toprule
\textbf{QLoRA Setting} & \textbf{Value} \\
\midrule
Quantization & 4-bit NormalFloat (NF4) \\
Compute dtype & bfloat16 \\
Double quantization & Yes \\
LoRA dropout & 0.05 \\
Target modules & q, k, v, o, gate, up, down proj \\
Bias & None \\
Task type & Causal LM \\
\midrule
\multicolumn{2}{l}{\textit{Per-model LoRA rank / alpha:}} \\
\midrule
SmolLM2-360M-Instruct & $r = 16$, $\alpha = 32$ \\
Qwen2.5-0.5B-Instruct & $r = 32$, $\alpha = 64$ \\
SmolLM3-3B & $r = 64$, $\alpha = 128$ \\
Qwen2.5-3B-Instruct & $r = 64$, $\alpha = 128$ \\
Llama-3.2-3B-Instruct & $r = 64$, $\alpha = 128$ \\
\bottomrule
\end{tabular}
\caption{QLoRA configuration. LoRA rank and alpha scale with model size to balance adapter capacity against overfitting risk.}
\label{tab:hyper_qlora}
\end{table}

\begin{table*}[t]
\centering
\small
\setlength{\tabcolsep}{5pt}
\renewcommand{\arraystretch}{1.15}
\begin{tabular}{l r r r r r r r}
\toprule
\textbf{Model} & \textbf{Params} & \textbf{Base} & \textbf{4-bit}$^\dagger$ & \textbf{Trainable} & \textbf{Adapter} & \textbf{Merged} & \textbf{Q4 GGUF}$^\dagger$ \\
 & & \textbf{(BF16)} & \textbf{(NF4)} & \textbf{params} & \textbf{(FP32)} & \textbf{(BF16)} & \\
\midrule
SmolLM2-360M-Instruct & 362M & 0.72\,GB & 0.26\,GB & 8.7M & 35\,MB & 0.72\,GB & 0.22\,GB \\
Qwen2.5-0.5B-Instruct & 494M & 0.99\,GB & 0.46\,GB & 17.6M & 70\,MB & 0.99\,GB & 0.30\,GB \\
SmolLM3-3B & 3.08B & 6.15\,GB & 1.98\,GB & 121M & 484\,MB & 6.15\,GB & 1.84\,GB \\
Qwen2.5-3B-Instruct & 3.09B & 6.17\,GB & 2.05\,GB & 120M & 479\,MB & 6.17\,GB & 1.85\,GB \\
Llama-3.2-3B-Instruct & 3.21B & 6.43\,GB & 2.24\,GB & 97.3M & 389\,MB & 6.43\,GB & 1.93\,GB \\
\bottomrule
\end{tabular}
\caption{Disaggregated model artifact sizes, replacing a single ambiguous ``checkpoint size'' figure. \textbf{Base (BF16)}, \textbf{Trainable params}, \textbf{Adapter (FP32)}, and \textbf{Merged (BF16)} are exact, measured from the released artifacts: the adapter is the saved \texttt{adapter\_model.safetensors} (re-serializing in bfloat16 would halve it), and the merged BF16 model equals the base size because LoRA deltas fold into existing weights without adding parameters. $^\dagger$The deployment-relevant 4-bit footprints, \textbf{4-bit (NF4)}, the QLoRA training/GPU backbone, and \textbf{Q4 GGUF}, the \texttt{Q4\_K\_M} CPU artifact, are estimates (NF4 from the quantized-linear plus retained-embedding split; GGUF at $\sim$0.6\,bytes/parameter), the same precision actually used at inference. Estimated single-example peak inference VRAM is approximately 0.6\,GB (360M), 0.9\,GB (0.5B), and 2.6--2.9\,GB (3B) at 4-bit on the RTX~4090; CPU-deployment peak RAM tracks the Q4 GGUF size plus a small ($<$0.5\,GB) runtime overhead.}
\label{tab:artifact_sizes}
\end{table*}

\begin{table}[h]
\centering
\small
\setlength{\tabcolsep}{6pt}
\renewcommand{\arraystretch}{1.15}
\begin{tabular}{ll}
\toprule
\textbf{Parameter} & \textbf{Value} \\
\midrule
Batch size & 8 \\
Max new tokens & 128 \\
Sampling & Yes (temperature $= 0.001$) \\
Top-$p$ & 1.0 \\
Repetition penalty & 1.0 \\
\bottomrule
\end{tabular}
\caption{Generation (inference) parameters used for all evaluation runs.}
\label{tab:hyper_generation}
\end{table}

\begin{table}[h]
\centering
\small
\setlength{\tabcolsep}{6pt}
\renewcommand{\arraystretch}{1.15}
\begin{tabular}{ll}
\toprule
\textbf{Parameter} & \textbf{Value} \\
\midrule
Base model & Llama-3.2-3B-Instruct \\
DAPT corpus & LitBank (train split) \\
DAPT learning rate & $2 \times 10^{-5}$ \\
DAPT max steps & 5{,}000 \\
DAPT batch size & 1 (grad.\ accum.\ 16) \\
DAPT warmup ratio & 0.03 \\
DAPT weight decay & 0.1 \\
DAPT sequence length & 1024 \\
Fine-tune regimes & LitTune, MixTune \\
Fine-tune shots & 0-shot only \\
\bottomrule
\end{tabular}
\caption{DAPT case study hyperparameters for the Llama-3.2-3B-Instruct-lit-dapt checkpoint. Fine-tuning uses the same settings as Table~\ref{tab:hyper_finetune}.}
\label{tab:hyper_dapt}
\end{table}

\section{Sequence Length and Truncation}
\label{app:truncation}
Two length caps apply in our pipeline. Training sequences are capped at \texttt{max\_seq\_length}${=}1{,}024$ tokens (Appendix~\ref{app:hyperparameters}), whereas inference inputs are capped only at each tokenizer's \texttt{model\_max\_length} ($8{,}192$ for SmolLM2-360M, up to $131{,}072$ for the others; the smallest underlying positional limit is $8{,}192$). Both caps truncate on the \emph{right}, and because every prompt places the final query and, in training, the appended gold label at the very end of the sequence, right-side truncation removes the gold label first and the query next; consequently, wherever training truncation occurs the gold label is dropped almost entirely (label-fully-cut rate ${\approx}$ truncation rate, e.g.\ $1.06\%$ of $1.10\%$ for PG-Fiction 2-shot).

Table~\ref{tab:truncation} summarizes input length and truncation per dataset for the highest-fertility tokenizer (SmolLM2-360M; the other four split text into fewer tokens). \textbf{Inference truncation is $0.00\%$ for every dataset, condition, and tokenizer}: the longest input anywhere is $2{,}500$ tokens, far below the smallest $8{,}192$-token context, so no evaluated prompt loses any context. Training truncation (the $1{,}024$ cap) is $0\%$ on all seven general-domain datasets and reaches at most ${\sim}1.1\%$ only on the longest literary dataset's 2-shot examples. This worst case is tokenizer-dependent: SmolLM2-360M truncates $1.105\%$ of PG-Fiction 2-shot training sequences, versus $0.737\%$ (SmolLM3-3B), $0.657\%$ (Llama-3.2-3B), and $0.485\%$ (the Qwen2.5 tokenizers). Measured directly on the actual fine-tuning mixtures rather than this test-split proxy, the same figure is $1.06\%$ (SmolLM2-360M), confirming the estimate is not an underestimate. REBEL's larger inference inputs reflect its large schema-enumerated label set rather than long documents, and remain far below the context cap. Per-tokenizer numbers for all $90$ cells are in \texttt{runs/evaluation/truncation\_stats.csv}, produced by \texttt{scripts/analyze\_truncation.py}.

\begin{table}[t]
\centering
\footnotesize
\setlength{\tabcolsep}{4pt}
\caption{Input length (tokens) and training-truncation rate per dataset, for the \textbf{SmolLM2-360M} tokenizer, typically the highest-fertility (most token-splitting) of the five, so these figures are a near-upper bound on input length across models. Med/Max are inference-input token counts; \%Trunc is the fraction of \emph{training} sequences exceeding the $1{,}024$-token cap. Inference truncation is $0.00\%$ for \emph{all} datasets, conditions, and tokenizers (global max input $2{,}500 < 8{,}192$, the smallest model context). Training truncation is $0\%$ everywhere except the three literary cells marked~$^{\dagger}$. Full per-tokenizer numbers: \texttt{runs/evaluation/truncation\_stats.csv}, via \texttt{scripts/analyze\_truncation.py}.}
\label{tab:truncation}
\begin{tabular}{l rrr rrr}
\toprule
& \multicolumn{3}{c}{0-shot} & \multicolumn{3}{c}{2-shot} \\
\cmidrule(lr){2-4}\cmidrule(lr){5-7}
Dataset & Med & Max & \%Trunc & Med & Max & \%Trunc \\
\midrule
TACRED & 426 & 673 & 0.00 & 641 & 984 & 0.00 \\
SemEval & 302 & 371 & 0.00 & 466 & 551 & 0.00 \\
CoNLL04 & 158 & 287 & 0.00 & 381 & 570 & 0.00 \\
NYT11 & 359 & 482 & 0.00 & 619 & 806 & 0.00 \\
GIDS & 245 & 458 & 0.00 & 596 & 1037 & 0.00 \\
Re-DocRED & 422 & 557 & 0.00 & 641 & 850 & 0.00 \\
REBEL & 1021 & 1190 & 0.00 & 1230 & 1638 & 0.00 \\
Biographical & 165 & 890 & 0.00 & 373 & 1589 & $0.013^{\dagger}$ \\
PG-Fiction & 709 & 1676 & $0.012^{\dagger}$ & 1152 & 2500 & $\mathbf{1.105}^{\dagger}$ \\
\bottomrule
\end{tabular}
\end{table}

\section{Frontier LLM Evaluation Protocol}
\label{app:frontier_prompts}

Frontier models (GPT-5.4 and Claude Sonnet~4.6) are evaluated via the OpenRouter API using the same prompt templates as the tuned SLMs. All frontier evaluations use 0-shot prompts with schema-enumerated prompting. The system and user messages are sent as a standard two-message chat completion request. All frontier results reported in this paper are scored on the same full test sets as the SLMs (no subsampling), with failed or empty API generations counted as errors, and use the same positive-class micro-F1 metric defined in Section~\ref{subsec:evaluation}.

\paragraph{Request configuration and reproducibility.}
Frontier requests use the OpenRouter model identifiers \texttt{openai/gpt-5.4}, \texttt{anthropic/claude-sonnet-4.6}, and \texttt{google/gemini-2.5-pro}, sent as two-message (system + user) chat-completion requests with \texttt{temperature}${=}0.0$, \texttt{top\_p}${=}1.0$, and \texttt{max\_tokens}${=}64$. We did \emph{not} set reasoning-effort, extended-thinking, or provider-routing parameters: requests therefore used each provider's default effort (documented in the providers' official model cards) and OpenRouter's default routing with fallback enabled, and we did not pin a backend provider. Because a 64-token output budget is far below what an extended-reasoning trace requires, these runs reflect the models' default, effectively non-reasoning configuration rather than an explicitly reasoning-maximized one. Transient failures (HTTP~429/502/503/504, timeouts, and API errors) were retried up to six times with exponential backoff at a request concurrency of~50; after retries, exactly one request failed across the entire frontier evaluation, one Claude Sonnet~4.6 call out of roughly $2.8\times10^{5}$ requests over the three models, and such failures are counted as errors in scoring. The frontier generations were collected in early April~2026. As a reproducibility limitation, we logged each response's content and token usage but \emph{not} the backend provider or model snapshot that OpenRouter returned per request; we release all collected frontier generations so that the outputs themselves remain auditable, and recommend that future evaluations additionally record the provider, model-snapshot, and request identifiers, and pin provider routing.

\paragraph{System prompt (schema-enumerated).}
\begin{verbatim}
You are a relation extraction system. Be concise
and direct. Output ONLY ONE relation type that
holds between the two mentioned entities. You MUST
choose exactly one label from this allowed set:
{allowed_labels}
Do not output any explanation, punctuation, or
extra text -- only the label.
\end{verbatim}

\noindent where \texttt{\{allowed\_labels\}} is replaced with the sorted, comma-separated set of gold relation labels for the target evaluation dataset.

\paragraph{User prompt.}
The user message is drawn directly from the pre-built \texttt{prompt\_0\_shot} column of each evaluation dataset, i.e., each instance is rendered with one of the ten instruction templates listed in Appendix~\ref{app:zero_shot}, assigned uniformly at random at dataset-construction time. Because the same pre-rendered column is used for both SLM and frontier evaluation, every frontier model receives exactly the same user prompt as the tuned SLMs for each instance. The only structural difference is that SLM few-shot prompts are unravelled into alternating user/assistant chat turns via the model's native chat template, whereas frontier models receive the prompt as a single user message (moot at 0-shot).

One representational difference concerns the catch-all class on TACRED and GIDS: in the SLM evaluation data the catch-all is an empty-string label (Appendix~\ref{app:dataset_stats}), whereas the frontier label sets render it as the named token \texttt{NA}, which the frontier models can emit explicitly. Each system is therefore scored against a catch-all representation it is able to produce.

\paragraph{Generation parameters.}
Frontier inference uses temperature~$= 0.0$, \texttt{max\_tokens}~$= 64$, and \texttt{top\_p}~$= 1.0$. These differ slightly from the SLM settings (temperature~$= 0.001$, \texttt{max\_new\_tokens}~$= 128$) but are functionally equivalent for single-label extraction, where the generated output is typically 1--3 tokens.

\section{Schema-Enumerated vs.\ Generic Prompting}
\label{app:constrained_vs_open}

Table~\ref{tab:constrained_vs_open_full} reports the per-dataset comparison between schema-enumerated and generic prompting across 164 matched-shot evaluations (excluding the two anomalous 0-shot configurations described in Section~4). Under schema-enumerated prompting, the system prompt enumerates the allowed label set for each evaluation dataset; under generic prompting, a generic system prompt is used without label enumeration. Both modes use the same fine-tuned checkpoints.

\begin{table*}[t]
\centering
\small
\setlength{\tabcolsep}{4pt}
\renewcommand{\arraystretch}{1.15}
\begin{tabularx}{\textwidth}{l >{\centering\arraybackslash}X >{\centering\arraybackslash}X >{\centering\arraybackslash}X >{\centering\arraybackslash}X >{\centering\arraybackslash}X >{\centering\arraybackslash}X >{\centering\arraybackslash}X >{\centering\arraybackslash}X >{\centering\arraybackslash}X c}
\toprule
& \multicolumn{7}{c}{\textbf{General}} & \multicolumn{2}{c}{\textbf{Literary}} & \\
\cmidrule(lr){2-8} \cmidrule(lr){9-10}
& TACRED & SemEval & CoNLL04 & NYT11 & GIDS & Re-DocRED & REBEL & Biogr. & PG-Fic. & \textbf{Mean} \\
\midrule
Schema-enum.\ F1 & 0.555 & 0.727 & 0.965 & 0.787 & 0.704 & 0.714 & 0.884 & 0.868 & 0.686 & 0.766 \\
Generic F1 & 0.609 & 0.784 & 0.951 & 0.808 & 0.833 & 0.716 & 0.886 & 0.895 & 0.702 & 0.798 \\
$\Delta$ (Generic $-$ Schema-enum.) & +0.053 & +0.057 & $-$0.015 & +0.022 & +0.129 & +0.002 & +0.002 & +0.027 & +0.015 & \textbf{+0.032} \\
\midrule
\multicolumn{11}{l}{\textit{Breakdown by model scale}} \\
\midrule
$\Delta$ sub-1B ($n = 72$) & +0.044 & +0.114 & $-$0.013 & +0.040 & +0.171 & +0.003 & +0.004 & +0.036 & +0.028 & +0.047 \\
$\Delta$ 3B ($n = 92$) & +0.061 & +0.011 & $-$0.016 & +0.007 & +0.096 & +0.001 & 0.000 & +0.021 & +0.006 & +0.021 \\
\bottomrule
\end{tabularx}
\caption{Schema-enumerated vs.\ generic prompting in \textbf{positive-class micro-F1} (no-relation class excluded), across 164 matched-shot evaluations (excluding the SmolLM3-3B MixTune 0-shot and Qwen2.5-3B GenTune 0-shot anomalies). Generic prompting outperforms schema-enumerated prompting on 8 of 9 datasets, with a mean improvement of $+3.2$ points (sub-billion $+4.7$, 3B $+2.1$). CoNLL04 is the sole exception ($-1.5$); GIDS shows the largest advantage ($+12.9$), whose tiny 4-relation schema with high label ambiguity benefits from the model's internalised schema rather than runtime enumeration. The improvement is \emph{not} a majority-class artifact: the all-class \emph{accuracy} gain is comparable ($+3.1$). Output well-formedness is essentially unchanged, schema-valid rate 0.882 (schema-enumerated) vs.\ 0.875 (generic); malformed-output rate $<$0.1\% under both, so label-set enumeration provides no formatting benefit that offsets its F1 cost.}
\label{tab:constrained_vs_open_full}
\end{table*}

\end{document}